\documentclass[wcp]{jmlr}
\usepackage[T1]{fontenc}
\usepackage{graphicx}
\usepackage{enumitem}
\usepackage{booktabs}
\usepackage[misc]{ifsym}
\usepackage{multirow}
\usepackage{amsmath}
\usepackage{amssymb}
\usepackage{graphicx}
\usepackage{wrapfig}
\usepackage{soul}
\usepackage{xcolor}
\usepackage{array}
\usepackage{float}
\usepackage{ulem}
\usepackage{caption}
\usepackage[ruled]{algorithm2e}

\newcolumntype{L}[1]{>{\raggedright\arraybackslash}m{#1}}

\usepackage{lineno}
\makeatletter
\let\Ginclude@graphics\@org@Ginclude@graphics 
\makeatother

\makeatletter

\newcommand{\@fs@ruled}{%
  \def\@fs@cfont{\bfseries}%
  \let\@fs@capt\floatc@ruled
  \def\@fs@pre{\hrule height 0.8pt depth 0pt \vskip 0.25em}%
  \def\@fs@post{\vskip 0.25em\hrule height 0.8pt depth 0pt \relax}%
  \def\@fs@mid{\vskip 0.2em}%
  \let\@fs@iftopcapt\iftrue
}
\newcommand{\fps@ruled}{htbp}       
\newcommand{\ftype@ruled}{4}        
\newcommand{\ext@ruled}{loa}        
\newcommand{\fnum@ruled}{\fname@ruled~\thealgorithm}
\newcommand{\fname@ruled}{Algorithm}
\floatstyle{ruled}
\renewcommand{\;}{
    \unskip\par
  \vskip\dimexpr\algomargin/4\relax
}
\restylefloat{algorithm}
\newcounter{algorithm} 
\renewcommand{\thealgorithm}{\arabic{algorithm}} 
\newcommand{\fnum@algorithm}{Algorithm~\thealgorithm}
\makeatother

\DeclareCaptionType{algobox}[Algorithm][List of Algorithms]

\begin{document}
\title[TAEGAN]{TAEGAN: Revisit GANs for Tabular Data Generation}





\newcounter{corrfn}


\author{
\Name{Jiayu Li}\thanks{Corresponding authors.}\setcounter{corrfn}{\value{footnote}}\Email{li.jiayu@nus.edu.sg}\\
\addr National University of Singapore, Singapore; Betterdata AI, Singapore
\AND
\Name{Zilong Zhao}\footnotemark[1]\Email{z.zhao@nus.edu.sg}\\
\addr National University of Singapore, Singapore; Betterdata AI, Singapore
\AND
\Name{Kevin Yee}\Email{kevin@betterdata.ai}\\
\addr Betterdata AI, Singapore
\AND
\Name{Uzair Javaid}\Email{uzair@betterdata.ai}\\
\addr Betterdata AI, Singapore
\AND
\Name{Biplab Sikdar}\Email{bsikdar@nus.edu.sg}\\
\addr National University of Singapore, Singapore
}

\makeatletter
\let \@jmlrpages \@empty
\makeatother

\maketitle
\begingroup
\renewcommand\thefootnote{}\footnotetext{This paper is accepted at ACML 2025.}
\addtocounter{footnote}{-1}
\endgroup
\SetKwComment{Comment}{/* }{ */}

\begin{abstract}
  Synthetic tabular data generation has gained significant attention for its potential in data augmentation and privacy-preserving data sharing. 
  While recent methods like diffusion and auto-regressive models (i.e., transformer) have advanced the field, generative adversarial networks (GANs) remain highly competitive due to their training efficiency and strong data generation capabilities.
  In this paper, we introduce Tabular Auto-Encoder Generative Adversarial Network (TAEGAN), a novel GAN-based framework that leverages a masked auto-encoder as the generator. TAEGAN is the first to incorporate self-supervised warmup training of generator into tabular GANs. It enhances GAN stability and exposes the generator to richer information beyond the discriminator's feedback. Additionally, we propose a novel sampling method tailored for imbalanced or skewed data and an improved loss function to better capture data distribution and correlations. We evaluate TAEGAN against seven state-of-the-art synthetic tabular data generation algorithms. Results from eight datasets show that TAEGAN outperforms all baselines on five datasets, achieving a 27\% overall utility boost over the best-performing baseline while maintaining a model size less than 5\% of the best-performing baseline model.
  Code is available at: \url{https://github.com/BetterdataLabs/taegan}.
\end{abstract}

\section{Introduction}
Synthetic data generation has gained tremendous attention due to its potential to provide an unlimited amount of data for data-hungry deep neural networks' training~\citep{aug,aug2}, and to unlock the use of a large amount of data with privacy and sensitivity concerns for sharing~\citep{gdpr}. The prevalence of tabular data modality in real-world applications, from governmental system records to commercial transaction records, also poses a great demand for synthetic tabular data.

\begin{figure}
    \centering
    \begin{minipage}[t]{0.6\textwidth}
    \vspace{-2em}
        \includegraphics[width=\linewidth]{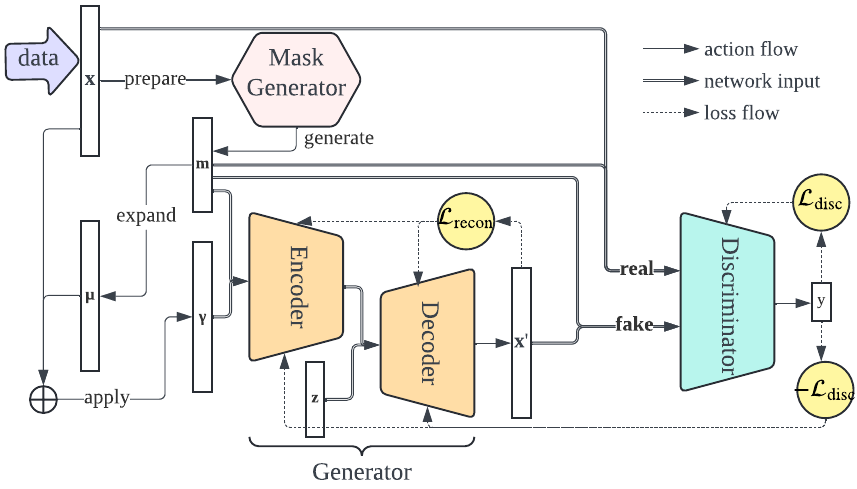}
        \caption{TAEGAN architecture. 
        The generator is a masked auto-encoder. The encoder processes masked real data to produce an encoded vector, which is passed to the decoder along with noise. The mask and decoder output are concatenated for discrimination. A reconstruction loss is applied alongside the GAN losses.
        }
        \label{fig:architecture}
    \end{minipage}
    \hfill
    \begin{minipage}[t]{0.35\textwidth}
    \vspace{-2em}
        \includegraphics[width=\linewidth,trim={7.5em 13em 9em 15.5em},clip]{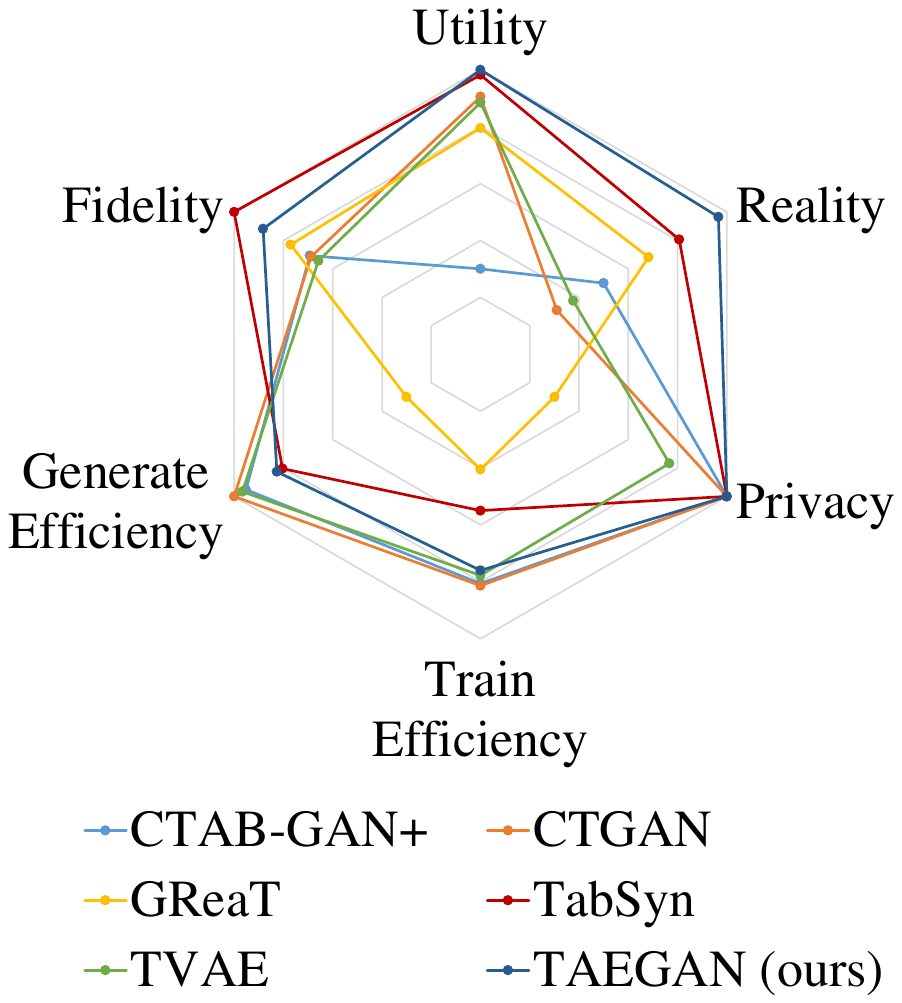}
        \vspace{-1.7em}
        \caption{Comparison of TAEGAN with
representative baseline models. TAEGAN demonstrates an overall advantage.
over baseline models.}
        \label{fig:overview}
    \end{minipage}
    \vspace{-1.5em}
\end{figure}


The research on synthetic tabular data generation using deep generative models initially focused on generative adversarial networks (GANs)~\citep{gan} and variational autoencoders (VAEs)~\citep{vae}. In recent years, diffusion models~\citep{diffusion} and auto-regressive models~\citep{gpt3} have gained significant traction due to their training stability and success in other modalities, such as images and text. However, GANs remain a competitive choice for tabular data generation, thanks to their unique advantages in training efficiency and generation speed~\citep{gan-eff,ar-eff}, as well as their strong empirical performance—particularly given concerns that vision and language models may be over-skilled for tabular tasks~\citep{revisit,bench}. In fact, a well-designed GAN can achieve comparable or even superior performance to diffusion and auto-regressive baselines while being far more efficient.

In this paper, we present such a design of tabular GAN: \textbf{T}abular \textbf{A}uto-\textbf{E}ncoder \textbf{G}enerative \textbf{A}dversarial \textbf{N}etwork (TAEGAN). Unlike the traditional design of tabular GANs where data are generated from noise or a sampled condition~\citep{ctgan,ctabgan}, TAEGAN generates data from a masked version of sampled real data and gradually removes the mask by iterative decoding~\citep{maskgit}. Its generator is a modified auto-encoder with two additional inputs: i) a mask to the encoder for masked modeling, and ii) a noise vector to the decoder for the completeness of GAN. 
Besides the standard GAN loss, a reconstruction loss is also added. 
The overall architecture of TAEGAN is shown in Fig.~\ref{fig:architecture}. 
TAEGAN’s generator is an independent auto-encoder, enabling self-supervised warmup training. The discriminator can be jointly trained in the warmup stage using the generator’s output.
This property not only improves the stability of GAN~\citep{unbalanced-gan}, which is a severe problem GANs have long suffered from, but also allows the generator to learn more about the real data by feedback beyond the discriminator's gradient update.

Besides the change in generative method and architecture, we also devise a novel data sampling approach during training to address the ubiquitous data imbalance and skewness problem. A novel loss component is also introduced to encourage the models to learn better data distributions and correlations. 

Experiments comparing 7 state-of-the-art (SOTA) synthetic tabular data generation models over 8 datasets show TAEGAN's stable advantage in terms of synthetic data quality and efficiency, as shown in Fig.~\ref{fig:overview}. 
Notably, although TAEGAN is less than 10\% the size of the best-performing baseline, it achieves a 27\% performance improvement on the most widely used evaluation metric—machine learning efficacy (MLE). Furthermore, TAEGAN outperforms all baseline models in MLE.
Additionally, thanks to the discriminator in the GAN framework, TAEGAN demonstrates a significant advantage in the reality of the synthetic data, as measured by the indistinguishability between real and synthetic data.

In summary, our major contributions include:
\begin{enumerate}[itemsep=0pt, parsep=0pt, topsep=0pt, partopsep=0pt, leftmargin=*]
    \item We introduce a novel tabular GAN framework, where the generator is a masked auto-encoder, enabling self-supervised warmup training. This approach enhances both training stability and the quality of the synthetic data.
    \item We propose a \textit{weight-matrix-based data sampling} method that addresses the common challenges of imbalance and skewness in tabular data during training. 
    \item We design a novel \textit{information loss} that is designed specifically for tabular generative models to learn better data distribution and correlation.
    \item 
    We propose TAEGAN, which leverages the above three innovations, achieving SOTA performance in synthetic data quality while offering significantly improved efficiency compared to the top-performing baseline models. 
\end{enumerate}


\section{Related Works}
\label{sec:related}
\subsection{Generating Synthetic Tabular Data}

When deep generative models were first introduced to the task of synthetic tabular data generation, GANs~\citep{gan} (e.g., CTGAN~\citep{ctgan}, CTAB-GAN~\citep{ctabgan}) and VAEs~\citep{vae} (e.g.,  TVAE~\citep{ctgan}, GOGGLE~\citep{goggle}) were the predominant generative paradigms. In particular, 
CTGAN and TVAE~\citep{ctgan} 
introduce tabular-specific feature engineering for these models and a data sampling method during training to address data imbalance.
One-hot encoding is applied for categorical features, and variational Gaussian mixture (VGM) model decomposition is applied to continuous features to naturally capture multimodal distributions. 
Training data is sampled following the logarithmic frequency based on real-data-derived condition vectors.
TAEGAN borrows or modifies these designs, with more details on necessary preliminaries seen in Section~\ref{sec:pre}.

Recently, the increasingly popular generative models for image and text generation--diffusion models~\citep{diffusion} and auto-regressive models~\citep{gpt2,gpt3}, begin to dominate the field of synthetic tabular data generation too (e.g., TabDDPM~\citep{tabddpm}, TabSyn~\citep{tabsyn}, and GReaT~\citep{great}), due to the high quality of the synthetic data. However, these models, especially the better-performing ones, typically require a transformer~\citep{transformer} network that is usually used for image or text tasks. For example, TabSyn~\citep{tabsyn} adapts a score-based diffusion model~\citep{score-diff,score-diff2} with a transformer architecture, which was initially designed for vision tasks, and GReaT~\citep{great} adapts a Distil-GPT2~\citep{gpt2}, which was initially designed for language tasks, with tabular-specific data encoding to generate synthetic tabular data.

However, while diffusion models and auto-regressive models generate synthetic data with impressive quality, the fact of using networks initially designed for images and text often implies a potential of over-fitting~\citep{revisit,bench}. Moreover, the efficiency of diffusion models and auto-regressive models are typically worse than GANs~\citep{gan-eff,ar-eff}. Therefore, there are still chances for a well-designed GAN specifically for tabular data to achieve comparable or better performance compared to these diffusion and auto-regressive models while being much more efficient.

\subsection{Masked Modeling and Its Applications in Tabular Data}

Masking is an important technique for self-supervised learning. 
Masked language modeling (MLM) has been a primary task for text pre-training~\citep{bert,roberta}, and masked image modeling (MIM) has played a similar role in image pre-training~\citep{mae,beit}. 
Data generation using masked modeling by gradually replacing masks with generated data has also gained great success in images~\citep{maskgit}.

In tabular data, masking is a natural method for imputation~\citep{remasker}. There are also generative models using masked modeling, such as TabMT~\citep{tabmt}, which combines masking with a transformer. However, it suffers from long training and sampling time due to its iterative sampling nature on a large model.

\section{Preliminaries and Notations}
\label{sec:pre}
In this section, we introduce the notations and preliminaries from prior works that are necessary to understand TAEGAN. Most notations are a generalized version of prior works that helps the understanding of this paper. 

\subsection{Tabular Data Generation Notations}
\label{sec:pre:pre}

\paragraph{Problem Formulation.}
Given a real table $\mathcal{T}$ with $M$ rows, $N_d$ categorical and $N_c$ continuous features ($N=N_d+N_c$ features in total), the objective is to generate synthetic $\mathcal{T}'$ of the same size that resembles the real one, i.e., $\mathcal{T}'\sim\mathcal{T}$.

\paragraph{Feature Engineering.} TAEGAN inherits feature engineering from CTGAN~\citep{ctgan}. Categorical features are one-hot encoded. Continuous features are decomposed using VGM to capture multi-modal distributions. Each value is represented by concatenating the one-hot encoded mode index (e.g., peak on the probability density graph) with the numeric value within that mode (following a normal distribution) (see details in \textit{mode-specific normalization} in \citep{ctgan}).

\paragraph{Tabular Notations.} Thus, formally, each feature of the table can be represented as a vector $\mathbf{x}_n,n\in\{1,2,\dots,N\}$. $\mathbf{x}_n$ consists of a discrete \textit{component} $\mathbf{d}_n$ (i.e., one-hot encoded category or mode for continuous feature) and a continuous \textit{component} $\mathbf{c}_n$ (i.e., empty for categorical and the numeric value within the mode for continuous feature), namely, $\mathbf{x}_n=\mathbf{d}_n\oplus\mathbf{c}_n$, where $\oplus$ denotes vector concatenation. 
A categorical feature has only a discrete \textit{component} and a continuous feature has a discrete and a continuous \textit{component}.
Then, let $|\cdot|$ be the number of dimensions of a vector, and let $D_n=|\mathbf{x}_n|,D_{dn}=|\mathbf{d}_n|,D_{cn}=|\mathbf{c}_n|$
($D_n=D_{dn}+D_{cn}$).
Thus, a row in $\mathcal{T}$ can be represented by the vector $\mathbf{x}=\bigoplus_{i=n}^N\mathbf{x}_n=\bigoplus_{n=1}^N\mathbf{d}_n\oplus\mathbf{c}_n$ with $C=N_d+2N_c$ (non-empty) \textit{components} and $D=\sum_{n=1}^ND_n$ dimensions. The ``feature engineering'' step in Fig.~\ref{fig:notation} shows an example of the representations.

\vspace{-0.5em}
\subsection{Conditional Vector}
\label{sec:pre:cond}
\begin{figure}[t]
    \centering
    \includegraphics[width=\linewidth]{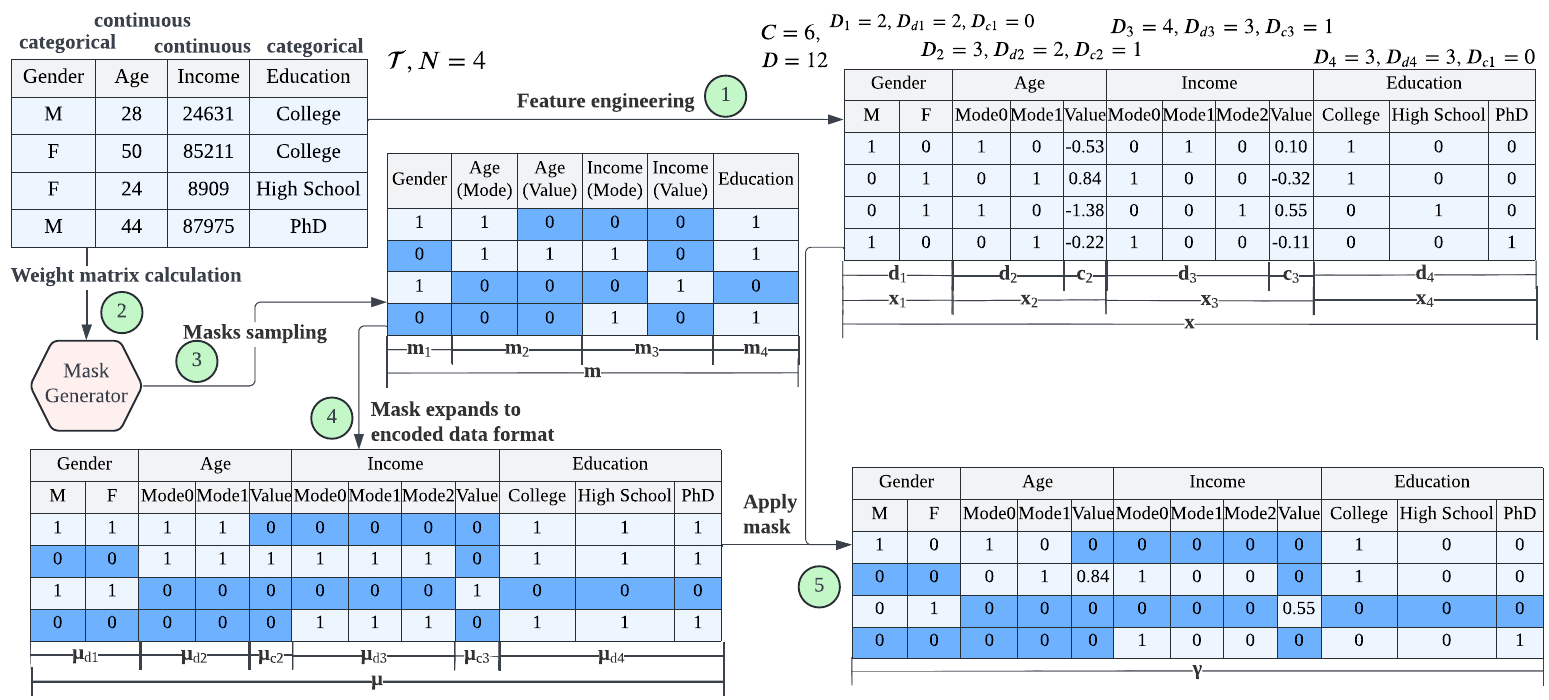}
    \vspace{-1.5em}
    \caption{Example of TAEGAN data processing and mask process. 
    }
    \label{fig:notation}
    \vspace{-2em}
\end{figure}
Prior work has found that tabular GANs perform better with a conditional generator~\citep{ctgan}. This requires a condition, we refer to as a \textit{conditional vector}, passed together with noise to guide the generation of each row.
The definition of \textit{conditional vector}s for tabular GANs has evolved since its first introduction~\citep{ctgan}. In this section, we introduce \ul{a new representation compatible with different prior works and our proposed model TAEGAN}.


We denote the \textit{conditional vector} as $\boldsymbol{\gamma}=\bigoplus_{n=1}^N\boldsymbol{\gamma}_{dn}\oplus\boldsymbol{\gamma}_{cn}$, where $\boldsymbol{\gamma}_{dn}$ and $\boldsymbol{\gamma}_{cn}$ correspond to $\mathbf{d}_n$ and $\mathbf{c}_n$ respectively. In some prior works, not all components can be part of the condition. For example, CTGAN allows only components from categorical features.
\textit{Components} ($\alpha$, can be $dn$ or $cn$) disallowed in the condition have a corresponding $\boldsymbol{\gamma}_{\alpha}=\boldsymbol{\phi},|\boldsymbol{\phi}|=0$. Let the number of allowed components in the condition be $\widehat{C}$, and $\mathbf{x}$ with allowed components only be denoted $\widehat{\mathbf{x}}$.
%
The \textit{conditional vector} is essentially a masked $\widehat{\mathbf{x}}$. Each \textit{component} is either fully masked or completely unmasked. Formally, let the \textit{mask indicator} show the masking status for each \textit{component} allowed in condition, represented as $\mathbf{m}=\bigoplus_{n=1}^N\mathbf{m}_{dn}\oplus\mathbf{m}_{cn}=(m_c)_{c=1}^{\widehat{C}}\in\{0,1\}^{\widehat{C}}$, where $|\mathbf{m}_{\alpha}|=0$ if $\boldsymbol{\gamma}_{\alpha}=\boldsymbol{\phi}$ and $|\mathbf{m}_{\alpha}|=0$ otherwise. In this paper, a mask value 1 means the value is maintained after masking.
The \textit{mask indicator} can be expanded by the dimensions of \textit{components} to create the \textit{data masks} as another binary vector $\boldsymbol{\mu}=\bigoplus_{i=1}^N\boldsymbol{\mu}_{di}\oplus\boldsymbol{\mu}_{ci}$, where $|\boldsymbol{\mu}_{\alpha}|=0$ if $\boldsymbol{\gamma}_{\alpha}=\boldsymbol{\phi}$ otherwise $|\boldsymbol{\mu}_{\alpha}|=D_{\alpha}$, and $\boldsymbol{\mu}_{dn},\boldsymbol{\mu}_{cn}\in\{\mathbf{0},\mathbf{1}\}$ (vectors with all 0s or with all 1s). Then, the \textit{conditional vector} is $\boldsymbol{\gamma} = \boldsymbol{\mu}^T \cdot \widehat{\mathbf{x}}$, which essentially means the masked data by the given \textit{masked indicator}. Appendix~\ref{app:taegan} contains a summary of these notations to aid understanding.

\begin{figure}[t]
    \centering
    \begin{minipage}[t]{0.55\linewidth}
    \begin{minipage}[t]{\linewidth}
    \vspace{-20em}
        \captionof{table}{Comparison of the definition of \textit{conditional vector}s in CT(GAN), CTAB(-GAN) and TAE(GAN), following the representation introduced in this paper.}
        \vspace{-1em}
        \label{tab:cond-comp}
        \resizebox{\linewidth}{!}{
        \begin{tabular}{L{0.18\linewidth}|ccc|L{0.4\linewidth}}
    \toprule
     & CT & CTAB & TAE & Remark \\
    \midrule
    $\mathbf{m}_{cn}$ & $=\boldsymbol{\phi}$ & $=\boldsymbol{\phi}$ & $\in\{0,1\}^1$ & cont. comp. allowed in TAE \\\hline
    $\mathbf{m}_{dn},$ $D_{cn}>0$ & $=\boldsymbol{\phi}$ & $\in\{0,1\}^1$ & $\in\{0,1\}^1$ & disc. comp. from cont. feat. allowed in CTAB and TAE \\\hline
    $\|\mathbf{m}\|_1,$ $|\mathbf{m}|>0$ & $=1$ & $=1$ & $\in\{0,1\}^{\widehat{C}}$ & no constraint for TAE \\
    \bottomrule
\end{tabular}
        }
    \end{minipage}
    \begin{minipage}[t]{\linewidth}
    \vspace{-7em}
        \includegraphics[width=\linewidth]{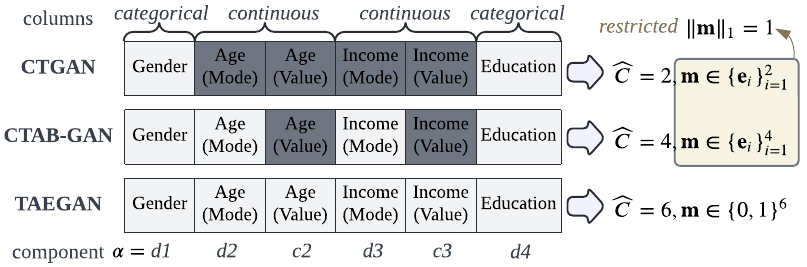}
        \vspace{-2em}
        \caption{An example of \textit{conditional vector}s of prior tabular GANs with TAEGAN.}
        \label{fig:cond-comp}
    \end{minipage}
    \hfill
    \end{minipage}
    \begin{minipage}[t]{0.44\linewidth}
        \includegraphics[width=\linewidth]{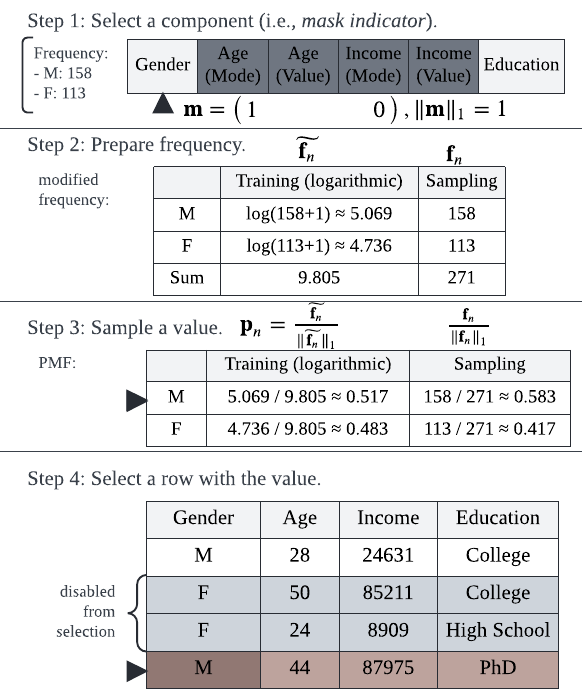}
        \vspace{-1.5em}
        \caption{Data sampling of CTGAN explained with the example. Black arrows indicate the selected ones.}
  \label{fig:ctgan}
    \end{minipage}
    \vspace{-2em}
\end{figure}

Many prior works also have the constraint that exactly one \textit{component} is maintained in the condition~\citep{ctgan,ctabgan}, formally, $\|\mathbf{m}\|_1=1$ if $|\mathbf{m}|>0$. In other words, $\mathbf{m}\in\{\mathbf{e}_i\}_{i=1}^{\widehat{C}}$ are one-hot vectors. Together with allowed \textit{components} in condition, \uline{constraints on $\boldsymbol{\gamma}$ of prior tabular GANs and TAEGAN are compared in Table~\ref{tab:cond-comp} and Fig.~\ref{fig:cond-comp}}. In TAEGAN, all \textit{components} are allowed in the condition, and hence $\mathbf{x}=\widehat{\mathbf{x}}$. In the rest of the paper, we use $\mathbf{x}$ for $\widehat{\mathbf{x}}$ without ambiguity. An example of TAEGAN \textit{conditional vector} construction is visualized in Fig.~\ref{fig:notation}.

\subsection{Data Sampling with Logarithmic Frequency during Training}
\label{sec:pre:sample}
\vspace{-0.5em}


During training, CTGAN samples data by logarithmically transformed frequency of a value of a randomly selected feature. The \textit{mask indicator} is first sampled, where only a categorical feature can remain unmasked. 
%
Then, with the \textit{mask indicator} fixed, a value in the selected categorical feature is sampled by logarithmically transformed frequency. Formally, let $\mathbf{f}_n=(f_{ni})_{i=1}^{D_{dn}}\in\mathbb{N}^{D_{dn}}$ be the frequency vector of the $n$-th (categorical) feature in real data. Let the logarithmically transformed frequency vector smoothed by 1 be $\widetilde{\mathbf{f}_n}=\log\left(\mathbf{f}_n+1\right)$. The probability mass function for sampling is thus a normalized version of $\widetilde{\mathbf{f}_n}$, i.e., $\mathbf{p}_n=\widetilde{\mathbf{f}_n}/\|\widetilde{\mathbf{f}_n}\|_1=(p_{ni})_{i=1}^{D_{dn}}$. The process is visualized in Fig.~\ref{fig:ctgan}

\section{TAEGAN: Optimized GANs for Tabular Data}
\label{sec:taegan}


\subsection{Overview and Architecture}
\label{sec:taegan:overview}

\paragraph{Architecture.} Recall the architecture of TAEGAN introduced in Fig.~\ref{fig:architecture}, \ul{the generator is a masked auto-encoder}. The encoder takes in the \textit{mask indicator} $\mathbf{m}$ and the masked data, i.e., \textit{conditional vector} $\boldsymbol{\gamma}=\boldsymbol{\mu}^T\cdot\mathbf{x}$. The encoded data with the noise $\mathbf{z}$ is concatenated and passed to the decoder, which generates synthetic data $\mathbf{x}'$. Since most tabular data are a mixture of discrete and continuous values, we modify the noise $\mathbf{z}$ to be a vector whose first half is discrete with values 0's and 1's, and second half is continuous following standard normal distribution.

\paragraph{Training.} Since the generator is essentially an auto-encoder, a reconstruction loss can be calculated between $\mathbf{x}$ and $\mathbf{x}'$, allowing the generator to be trained without the discriminator, which we refer to as the warmup of GAN training.
While the generator is trained in the warmup stage, the discriminator can be trained jointly but without direct interaction with the training of the generator in terms of gradients. Consequently, \ul{both the generator and discriminator can be warmed up independently, conceptually similar to a pre-training on the provided dataset}. This allows a better initialization of both networks before the adversarial training of GAN starts, enabling improved stability during adversarial training~\citep{unbalanced-gan}. The overall training process is described in Algorithm~\ref{alg:train}, where details of some steps are elaborated in subsequent sections.

\begin{figure}[t]
    \centering
    \begin{minipage}{0.92\linewidth}
        \begin{algorithm}[H]
            \footnotesize
            \LinesNumbered
            \SetAlgoNoEnd
            \SetKwFor{RepeatTimes}{repeat }{ times:}{}
            \captionsetup{type=algobox}
\captionof{algobox}{TAEGAN Training Algorithm}
            \label{alg:train}
            \KwData{Training table $\mathcal{T}$, warmup \& main epochs $E_p,E$}
            \KwResult{Generator parameters $\boldsymbol{\theta}_G$}
            Preprocess table $\mathcal{T}$ by one-hot encoding and VGM decomposition \;
            Calculate $\mathcal{W}$ to prepare for sampling\label{line:weight} \;
            Initialize network parameters $\boldsymbol{\theta}_G,\boldsymbol{\theta}_D$\;
            \For{$e$ \textbf{in} $1,2,\dots,E_p+E$}{
                \RepeatTimes{$M$}{
                  $\mathbf{x},\mathbf{m},\boldsymbol{\mu},\boldsymbol{\gamma}\gets$ sampled data and masks\label{line:dm-sample} \Comment*[r]{Based on $\mathcal{W}$}
                  Generate $\mathbf{x}'\gets G(\mathbf{m},\boldsymbol{\gamma},\mathbf{z})$ based on noise $\mathbf{z}$ from latent space\;
                  Update $\boldsymbol{\theta}_G$ based on reconstruction loss, i.e., $\mathcal{L}_{\text{recon}}(\mathbf{x}',\mathbf{x},\mathbf{m})$\;
                  Discriminate the data $y_r\gets D(\mathbf{m},\mathbf{x}),y_f\gets D(\mathbf{m},\mathbf{x}')$\;
                  Update $\boldsymbol{\theta}_D$ based on discrimination, i.e., $\mathcal{L}_{\text{disc}}(y_r,y_f)$\;
                  \If{$e>E_p$}{
                    \Comment{Do the adversarial generation after warmup training}        
                    Update $\boldsymbol{\theta}_G$ based on incorrect discrimination, i.e., $-\mathcal{L}_{\text{disc}}(y_r,y_f)$
                  }
                }
            }
        \end{algorithm}
    \end{minipage}
    \label{fig:taegan_train}
\end{figure}

\paragraph{Generation.} In TAEGAN, the masked auto-encoder design of the generator allows the mask $\mathbf{m}$ to be arbitrary as long as it falls in the range of $\{0,1\}^C$, in contrast to CTGAN~\citep{ctgan}'s mask with several constraints (recall Section~\ref{sec:pre:cond}). Consequently, \ul{a single row can be generated in multiple steps by iterative decoding}, in contrast to CTGAN's one-time generation from the \textit{conditional vector}. In the extreme case, one component is generated per step. The first component (step) can be directly taken from real data, which exposes only one component of one row, generally considered safe without privacy concerns as it contains a similar amount of information to the \textit{conditional vector} of CTGAN. Although each time all dimensions are generated, we select only one additional from the input and proceed to the next iteration.
The full process is visualized in Fig.~\ref{fig:gen}, and details to be found in Appendix~\ref{app:taegan}.

\begin{figure}[t]
    \vspace{-1em}
    \centering
    \includegraphics[width=\linewidth]{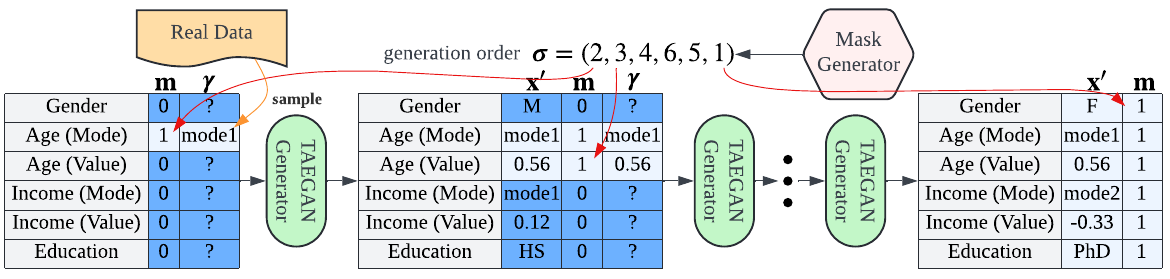}
    \caption{Generation process of TAEGAN. Components are generated one by one following a sampled order. The first component is sampled from real data.}
    \label{fig:gen}
    \vspace{-1.5em}
\end{figure}

\begin{figure}
\hfill
    \begin{minipage}[t]{0.49\linewidth}
        \input{algos/weight}
    \end{minipage}
    \hfill
    \begin{minipage}[t]{0.44\linewidth}
        \input{algos/sample}
    \end{minipage}
    \vspace{-1.5em}
\end{figure}
\subsection{Data Sampling during Training}
\label{sec:taegan:sample}

Now that all constraints on $\mathbf{m}$ are relaxed (recall Section~\ref{sec:pre:cond}), the sampling of \textit{mask indicator} and \textit{conditional vector} are adjusted accordingly, to allow sampling based on conditions that are multivariate and potentially continuous.

\paragraph{Sampling the \textit{Mask Indicator} $\mathbf{m}$.} The \textit{mask indicator} is sampled by first sampling a number of components (value of $\|\mathbf{m}\|_1$) from $\{1,2,\dots,C\}$, then randomly sampling this number of components. \textit{Mask indicators} activating zero components (i.e., full mask) skipped during training because they are not used during generation (recall Algorithm~\ref{alg:sample}). Also, note that it is generally harder to generate data with less information, so we sample smaller $\|\mathbf{m}\|_1$ with a higher probability mass than larger ones. In this paper, we use the normalized inverse of the values as the probability mass, namely, $\forall X\in\{1,2,\dots,C\}$,
\vspace{-0.5em}
\begin{equation}
\small
    \label{eq:n-mask}
    \mathbb{P}(\|\mathbf{m}\|_1=X)=\frac{1/X}{\sum_{i=1}^C1/X}
\end{equation}

\paragraph{Sampling the \textit{Conditional Vector} $\boldsymbol{\gamma}$ and Data.} \vspace{-0.3em} Given a \textit{mask indicator} without the constraints for CTGAN, the probability mass or density function of the corresponding \textit{conditional vector} $\boldsymbol{\mu}$ cannot be computed by the feature-wise distribution as CTGAN does. Instead of sampling values based on the \textit{mask indicator}, we \ul{sample real data rows and then apply the masks on them}.

\ul{To calculate the probability for each row given the \textit{mask indicator}, we compute the \textit{weight matrix} $\mathcal{W}\in[0,1]^{M\times C}$} as combination of column vectors $\mathbf{w}_i\in[0,1]^M,i\in\{1,2,\dots,C\}$ with $\|\mathbf{w}_i\|_1=1$, which is the probability mass function over all rows in $\mathcal{T}$ when the $i$-th component is selected. 
Given the same \textit{mask indicator}, rows with the same values after masking are sampled uniformly.
Therefore, $w_{ij}=p_{nk}/f_{nk}$ if the $i$-th component is the discrete component of the $n$-th feature, and $\mathbf{d}_n=\mathbf{e}_k$ on the $i$-th row of $\mathcal{T}$. Continuous components' weights are calculated similarly by discretizing the values into bins so that frequencies and probability mass based on logarithmically transformed frequency can also be computed.
The full process of \textit{weight matrix} computation is described in Algorithm~\ref{alg:weight}, and is used for Line~\ref{line:weight} of Algorithm~\ref{alg:train}.

Finally, the dot product of $\mathcal{W}$ and normalized \textit{mask indicator} $\mathbf{m}$, i.e., $\boldsymbol{\rho}=\mathcal{W}\cdot\left(\mathbf{m}/\|\mathbf{m}\|_1\right)$ indicate the probability mass function of each row to be sampled under this $\mathbf{m}$ by the multiplication law of probabilities.
The full process is shown in Algorithm~\ref{alg:log-sample}, and used for Line~\ref{line:dm-sample} of Algorithm~\ref{alg:train}. 
Appendix~\ref{app:taegan} provides an illustrative example to assist understanding.


\begin{proposition}
    \label{thm:log-ctgan}
    The probability density/mass $\mathbb{P}(\boldsymbol{\gamma}|\mathbf{m})$ is identical from CTGAN~\citep{ctgan} (recall Section~\ref{sec:pre:sample}) and from Algorithm~\ref{alg:log-sample}, if $\mathbf{m}$ is allowed in both.
\end{proposition}

Proposition~\ref{thm:log-ctgan} showcases the consistency of TAEGAN's design with prior works (e.g., CTGAN), but TAEGAN is a generalized version. The proof is seen in Appendix~\ref{app:proof}.

\subsection{Loss Functions}

\paragraph{Reconstruction loss.} Reconstruction loss can be computed on each component separately. We use cross-entropy for discrete and smooth L1 loss~\citep{smoothl1} for continuous components. We denote the loss for the $i$-th component as $\ell_i(\mathbf{x}',\mathbf{x})\in\mathbb{R}$. The overall reconstruction loss is a weighted sum of losses on all components. We apply different weights on masked (known) and unmasked (unknown) components. Moreover, note that it is generally harder to reconstruct data from less known components because of the more diverse potential given limited known components, and easier from more known components as the reconstruction task eventually degenerates to classification or regression with increasing known components. Therefore, we adjust the weights on unknown components such that lower weights are applied when fewer components are known (smaller $\|\mathbf{m}\|_1$). We define a range of weights of unknown components as opposed to known components: $[\lambda_1,\lambda_2]\subseteq[0,1]$, then the reconstruction loss on one sample is:
\vspace{-0.5em}
\begin{equation}
\small
    \mathcal{L}_{\text{recon}}(\mathbf{x}',\mathbf{x},\mathbf{m})=\sum_{i=1}^C\left(m_i+(1-m_i)\left(\frac{\|\mathbf{m}\|_1\cdot(\lambda_2-\lambda_1)}{C}+\lambda_1\right)\right)\cdot\ell_i(\mathbf{x}',\mathbf{x})
\end{equation}

\vspace{-0.3em}
\ul{The reconstruction loss is used to update the generator's gradients in both the warmup and main training stages}. During warmup, this is the main loss function for the generator. During adversarial training, the reconstruction loss works in place of the auxiliary classification loss~\citep{clf-loss} that is also applied in some baseline tabular GANs~\citep{ctabgan}, as the reconstruction based on arbitrary masks is essentially a generalized version of the auxiliary classification task, and the generator network can be reused for classification to reduce the model size.

\paragraph{GAN loss.} For better GAN training stability, we adopt WGAN-GP~\citep{wgan-gp} following CTAB-GAN~\citep{ctabgan,ctabganp}, and PacGAN~\citep{pacgan} framework (i.e., the discriminator takes in ``pacs'' of, or a number of rows of, data) following CTGAN~\citep{ctgan}. However, for simplicity of higher-level understanding, we abstract them as $\mathcal{L}_{\text{disc}}(y_r,y_f)$ like a classical GAN~\citep{gan} in Algorithm~\ref{alg:train}.

\paragraph{Information loss.} To capture a better data distribution, we adopt the additional information loss for the generator following CTAB-GAN~\citep{ctabgan}, which calculates the mean and standard deviations of the vector before the last layer of the discriminator of real and synthetic data respectively on every batch, and the loss is their differences. The information loss can be further enriched by direct calculations on $\mathbf{x}$ and $\mathbf{x}'$ utilizing properties of the feature-engineered data, which consists of one-hot discrete components and normal-distributed continuous components (recall Section~\ref{sec:pre:pre}). Similarly to the original information loss, the losses are calculated by the differences of the aggregated values per batch between real and synthetic data. \ul{We design two additional loss components for marginal distribution and correlation respectively}:
\begin{itemize}[itemsep=0pt, parsep=0pt, topsep=0pt, partopsep=0pt,leftmargin=*]
    \item \textbf{Mean loss for marginal distribution.} The aggregated value for \textit{mean loss} is the mean for each dimension in the batch. In particular, the mean of one-hot vectors expresses exactly the probability mass function. 
    \item \textbf{Interaction loss for correlation.} The \textit{outer product} of two one-hot vectors of arbitrary dimensions (can be different dimensions) expresses a unique combination of the two input vectors, and the flattened vector is also one-hot. The \textit{outer product} of a one-hot vector with a one-dimensional scalar also expresses a unique combination of the two input values if the result is not $\mathbf{0}$. Therefore, the correlation can be captured utilizing \textit{outer product}s. The aggregated values for \textit{interaction loss} are the mean and standard deviation of the flattened \textit{outer product} $\mathbf{x}\otimes\mathbf{x}$ (supposing $\mathbf{x}$ is feature-engineered). In particular, dimensions in the \textit{outer product} corresponding to impossible value combinations of two discrete components will always be 0, which can be fully captured by the aggregated values.
\end{itemize} 

In summary, let $\mathbf{x}^{[k]}$ represent the $k$-th sample in a batch of size $B$, let $\Delta_{\text{mean}}(f(\mathbf{x}'),f(\mathbf{x}))$ where $f$ is an aggregation function on $\mathbf{x}$ and $\mathbf{x}'$ denote the difference of mean aggregated values, i.e., $\|\text{\textsc{Mean}}_{k=1}^B(f(\mathbf{x}'^{[k]}))-\text{\textsc{Mean}}_{k=1}^B(f(\mathbf{x}^{[k]}))\|_1$, and let $\Delta_{\text{std}}(f(\mathbf{x}'),f(\mathbf{x}))$ be the difference of standard deviation similarly. Then, the modified information loss of a batch is
\vspace{-0.5em}
\begin{multline}
\small
    \mathcal{L}_{\text{info}}(\mathbf{x}'^{[1..B]},\mathbf{x}^{[1..B]})=
    \alpha_1\cdot(\Delta_{\text{mean}}(\widetilde{D}(\mathbf{x}'),\widetilde{D}(\mathbf{x}))+\Delta_{\text{std}}(\widetilde{D}(\mathbf{x}'),\widetilde{D}(\mathbf{x})))\\
    +\alpha_2\cdot\Delta_{\text{mean}}(\mathbf{x}',\mathbf{x})
    +\alpha_3\cdot(\Delta_{\text{mean}}(\mathbf{x}'\otimes\mathbf{x}',\mathbf{x}\otimes\mathbf{x})+\Delta_{\text{std}}(\mathbf{x}'\otimes\mathbf{x}',\mathbf{x}\otimes\mathbf{x}))
\end{multline}
where $\widetilde{D}$ represents the discriminator before the last layer, and $\alpha_1,\alpha_2,\alpha_3$ are the weights on the original information, mean, and interaction losses respectively.

\paragraph{Overall Objective.} Let $p_{\mathbf{z}}$ be the probability distribution of the mixed-discrete-continuous noise $\mathbf{z}$, $p_{\mathbf{m}}$ is the probability distribution of $\mathbf{m}$ and $p_{\mathbf{x}|\mathbf{m}}$ is the probability distribution computed based on $\mathcal{W}$ given $\mathbf{m}$ as described in Section~\ref{sec:taegan:sample}.
Combining these losses, we have the objective of TAEGAN as 
\vspace{-0.5em}
\begin{multline}
\small
    \min_G\max_D\mathbb{E}_{\mathbf{z}\sim p_{\mathbf{z}},\mathbf{m}\sim p_{\mathbf{m}}}\mathbb{E}_{\mathbf{x}\sim p_{\mathbf{x}|\mathbf{m}}}\\\left(\mathcal{L}_{\text{recon}}(G(\mathbf{z},\boldsymbol{\gamma}),\mathbf{x},\mathbf{m})-\mathcal{L}_{\text{disc}}\left(D(\mathbf{m},\mathbf{x}),D(\mathbf{m},G(\mathbf{z},\boldsymbol{\gamma})\right)+\mathcal{L}_{\text{info}}(G(\mathbf{z},\boldsymbol{\gamma}),\mathbf{x})\right)
\end{multline}

\section{Experiments}
\label{sec:exp}
\subsection{Experiment Setup}

\paragraph{Datasets.} We conduct experiments on 8 datasets from OpenML~\citep{openml}: \texttt{adult}, \texttt{bank-marketing}, \texttt{breast-w}, \texttt{credit-g}, \texttt{diabetes}, \texttt{iris}, \texttt{qsar-biodeg}, and \texttt{wdbc}. Details can be found in Appendix~\ref{app:setup:datasets}.

\paragraph{Baselines.} We compare the performance of TAEGAN with several representative tabular generative models as baselines, including non-neural-network models, GANs, VAEs, diffusion models, and auto-regressive models: ARF~\citep{arf}, CTAB-GAN+ (CTAB+)~\citep{ctabganp}, CTGAN~\citep{ctgan}, TabDDPM (TDDPM)~\citep{tabddpm}, GReaT~\citep{great}, TabSyn~\citep{tabsyn}, and TVAE~\citep{ctgan}. Implementation details can be found in Appendix~\ref{app:setup:baselines}.

\paragraph{TAEGAN Configuration.} We use multi-layer perceptrons (MLP) as both the generator and discriminator. The networks are warmed up for 50 epochs capped at 500 steps, and then trained with adversarial losses for 300 epochs capped at 3000 steps, with a batch size of 3000 and learning rate of $2\times10^{-4}$. More details can be found in Appendix~\ref{app:setup:taegan}.
              
\paragraph{Testbed.} Experiments were conducted on a Ubuntu 22.04 server with an Intel i9-13900K CPU, 125 GB of RAM, and an NVIDIA RTX 4090 GPU.

\paragraph{Metrics.} We evaluate tabular generative models using utility, reality, fidelity, privacy, and efficiency, which are standard metrics in tabular data generation~\citep{ctgan,ctabganp,great,tabsyn,sdmetrics}. All results reported are based on 3 runs of each experiment. Detailed implementation of the metrics can be found in Appendix~\ref{app:setup:mle}-\ref{app:setup:efficiency}.
\begin{itemize}[itemsep=0pt, parsep=0pt, topsep=0pt, partopsep=0pt,leftmargin=*]
    \item \textbf{Utility} is usually tied to a downstream task (e.g., classification). The established machine learning efficacy (MLE) metric~\citep{ctgan} has synthetic data generated from a real training set tested on a hold-out real test set. We use 3 downstream models: logistic regression (LG), random forest (RF)~\citep{rf}, and XGBoost (XGB)~\citep{xgboost}. \ul{Better generative models yield higher weighted AUC ROC MLE results.}
    \item \textbf{Reality} means the indistinguishability of real and synthetic data~\citep{sdmetrics}. The same 3 ML models as utility are used for this binary classification task. AUC ROC is reported, with \ul{a lower score (above 0.5) means a higher difficulty to differentiate real from synthetic, hence better utility}.
    \item \textbf{Fidelity} shows the cosmetic discrepancy between the real and synthetic data in marginal distribution (MD, i.e., feature-wise) and correlation (CR, i.e., pair-wise)~\citep{sdmetrics,tabsyn}. \ul{Higher similarity between real and synthetic data means better data fidelity.}
    \item \textbf{Privacy} is crucial when the synthetic data is used for privacy-preserving data sharing. We use the distance to the closest record (DCR)~\citep{ctabgan} to evaluate the privacy. The DCR from synthetic data and from hold-out real (test) data to the real training data are compared. Privacy is preserved if the latter DCRs are no smaller than the former, tested by Mann-Whitney U Test~\citep{mwu}. \ul{Privacy is at risk when $p$-values are smaller than 0.05.}
    \item \textbf{Efficiency} is demonstrated by model sizes and computation time. \ul{Under comparable performance, smaller models and faster computation are preferred.}
\end{itemize}

\subsection{Comparison to Baselines}
\label{sec:exp:exp}

\begin{table}[t]
    \centering
    \caption{MLE performance summary. The first row is the overall average score. ``RE'' stands for the average relative error with the score produced by real data. The next 3 rows are the average scores on different downstream models. 
    The best scores are in bold and underlined, and the second-best scores are in bold.}
    \label{tab:mle}
    \vspace{-0.5em}
    \setlength{\tabcolsep}{10pt}
    \resizebox{\linewidth}{!}{
    \begin{tabular}{l|c|ccccccc|c}
\toprule
 & real & ARF & CTAB+ & CTGAN & TDDPM & GReaT & TabSyn & TVAE & TAEGAN \\
\midrule
 All ($\uparrow$) & 0.930 & 0.907 & 0.750 & 0.895 & 0.850 & 0.868 & \textbf{0.913} & 0.890 & \textbf{\underline{0.917}} \\
 RE ($\downarrow$) & - & 2.632\% & 19.217\% & 3.873\% & 8.604\% & 6.897\% & \textbf{1.951\%} & 4.511\% & \underline{\textbf{1.422\%}}\\
 LG ($\uparrow$) & 0.928 & 0.909 & 0.749 & 0.896 & 0.880 & 0.870 & \textbf{\underline{0.915}} & 0.882 & \textbf{0.914} \\
 RF ($\uparrow$) & 0.930 & 0.903 & 0.755 & 0.892 & 0.866 & 0.866 & \textbf{0.912} & 0.892 & \textbf{\underline{0.918}} \\
 XGB ($\uparrow$) & 0.932 & 0.908 & 0.745 & 0.896 & 0.803 & 0.869 & \textbf{0.912} & 0.895 & \textbf{\underline{0.919}} \\
\bottomrule
\end{tabular}

    }
    \vspace{-1em}
\end{table}

\begin{table}[t]
    \centering
    \caption{RSD performance summary. The first row is the average score of all experiments, and the other rows are the averages using different models. The best scores are in bold and underlined, and the second best scores are in bold.}
    \label{tab:rsd}
    \vspace{-0.5em}
    \setlength{\tabcolsep}{13pt}
    \resizebox{\linewidth}{!}{
    \begin{tabular}{l|ccccccc|c}
\toprule
 & ARF & CTAB+ & CTGAN & TDDPM & GReaT & TabSyn & TVAE & TAEGAN \\
\midrule
All ($\downarrow$) & 0.771 & 0.834 & 0.902 & 0.818 & 0.768 & \textbf{0.723} & 0.878 & \textbf{\underline{0.666}} \\
LG ($\downarrow$) & \textbf{0.610} & 0.711 & 0.829 & 0.745 & 0.696 & 0.636 & 0.724 & \textbf{\underline{0.588}} \\
RF ($\downarrow$) & 0.843 & 0.890 & 0.936 & 0.850 & 0.804 & \textbf{0.743} & 0.948 & \textbf{\underline{0.676}} \\
XGB ($\downarrow$)& 0.860 & 0.900 & 0.941 & 0.857 & 0.803 & \textbf{0.789} & 0.963 & \textbf{\underline{0.732}} \\
\bottomrule
\end{tabular}

    }
    \vspace{-1em}
\end{table}

\paragraph{Utility: MLE.} Table~\ref{tab:mle} summarizes the MLE performances of different generators. TAEGAN demonstrates a better overall performance than all baseline models, and wins over GAN baselines by a significant margin. In particular, TAEGAN outperforms the best-performing baseline's relative error compared to real by a factor of 27\%. Comparing the performances on different downstream models, TAEGAN shows a clearer advantage on more complex ones (RF, XGB) than simpler ones (LG). This also implies the effect of a masked auto-encoder and the interaction loss in capturing complex data correlations. 

\begin{table}[t]
    \centering
    \caption{Average MD and DC scores over all 8 datasets. The best scores are in bold and underline, and the second best scores are in bold. The best score among GANs (${}^*$) are shown with superscript ${}^*$.}
    \label{tab:fidelity}
    \vspace{-0.5em}
    \setlength{\tabcolsep}{10pt}
    \resizebox{\linewidth}{!}{
    \begin{tabular}{l|ccccccc|c}
\toprule
 & ARF & CTAB+${}^*$ & CTGAN${}^*$ & TDDPM & GReaT & TabSyn & TVAE & TAEGAN${}^*$ \\
\midrule
MD ($\uparrow$) & \textbf{0.925} & 0.898 & 0.852 & 0.751 & 0.881 & \textbf{\underline{0.931}} & 0.860 & $0.906^*$\\
CR ($\uparrow$) & \textbf{0.904} & 0.825 & 0.857 & 0.783 & 0.867 & \textbf{\underline{0.919}} & 0.839 & $0.892^*$\\
\bottomrule
\end{tabular}

    }
    \vspace{-1em}
\end{table}

\begin{table}[t]
    \centering
    \caption{Number of datasets (among 8 in total) where the synthetic data has a risk of privacy leakage.}
    \label{tab:dcr}
    \vspace{-0.5em}
    \setlength{\tabcolsep}{15pt}
    \resizebox{\linewidth}{!}{
    \begin{tabular}{ccccccc|c}
\toprule
ARF & CTAB+ & CTGAN & TDDPM & GReaT & TabSyn & TVAE & TAEGAN \\
\midrule
0 & 0 & 0 & 1 & 3 & 0 & 1 & 0\\
\bottomrule
\end{tabular}

    }
    \vspace{-1em}
\end{table}

\paragraph{Reality: RSD.} Table~\ref{tab:rsd} summarizes the RSD performances of different generators. TAEGAN demonstrates a clear advantage over all baseline models. GANs outperforming diffusion models and auto-regressive models on RSD is unsurprising due to the inherent presence of a discriminator, which is essentially an RSD task using a neural network. However, the fact that baseline GANs cannot outperform certain models in other generation paradigms shows that prior tabular GAN models have left significant room for further optimization.

\paragraph{Fidelity: MD and CR.} Table~\ref{tab:fidelity} shows the average MD and DC scores.
Some baseline models outperform TAEGAN, likely due to the logarithmic frequency during training in TAEGAN, which can slightly distort the distribution. 
All GAN frameworks experimented adopt a certain training with logarithmic frequency, and TAEGAN is clearly the best GAN model, demonstrating the effectiveness of the information loss construction.

\paragraph{Privacy: DCR.} Table~\ref{tab:dcr} shows the number of datasets where DCRs suggest a risk of privacy leakage. TAEGAN do not pose outstanding privacy risks. 

\begin{table}[t]
    \centering
    \caption{Average training and generation times (in seconds) across all 8 datasets.
    Generation means generating the same amount of data as the training set.}
    \label{tab:time}
    \vspace{-0.5em}
    \setlength{\tabcolsep}{12pt}
    \resizebox{\linewidth}{!}{
    \begin{tabular}{l|ccccccc|c}
\toprule
 & ARF & CTAB+ & CTGAN & TDDPM & GReaT & TabSyn & TVAE & TAEGAN \\
\midrule
Training & 51.933 & 187.517 & 178.590 & 67.330 & 2660.705 & 1023.474 & 226.858 & 254.823 \\
Generation & 10.147 & 0.302 & 0.208 & 15.776 & 47.157 & 0.952 & 0.270 & 0.798 \\
\bottomrule
\end{tabular}

    }
    \vspace{-0.5em}
\end{table}
\begin{figure}[t]
    \centering
    \begin{minipage}{0.49\linewidth}
        \centering
        \includegraphics[width=\linewidth]{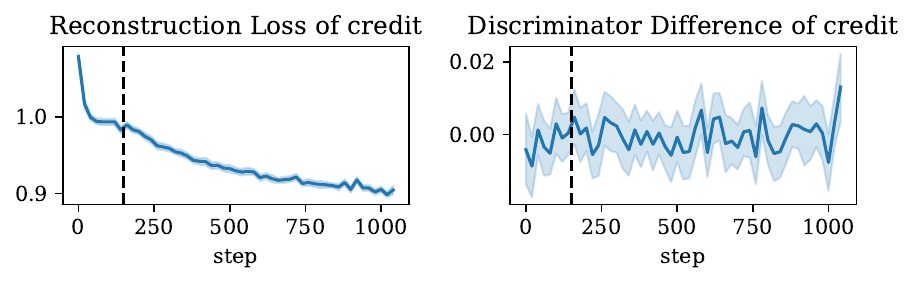}
    \end{minipage}
    \begin{minipage}{0.50\linewidth}
        \centering
        \includegraphics[width=\linewidth]{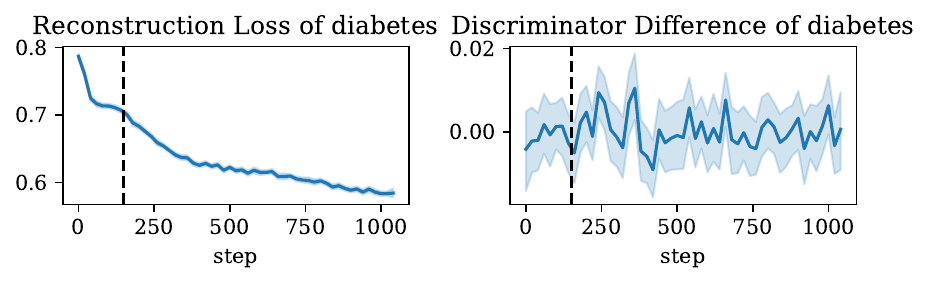}
    \end{minipage}
    \vspace{-1em}
    \caption{Dynamic of losses on \texttt{credit} and \texttt{diabetes}. ``Reconstruction Loss'' is in log10 scale for visibility. ``Discriminator Difference'' means $y_r-y_f$. The black vertical line is the divider between the warmup and main training stages.}
    \label{fig:loss}
    \vspace{-1.5em}
\end{figure}

\paragraph{Efficiency: Computation Time.} Table~\ref{tab:time} shows the average training and generation time over all 8 datasets of different models. The state-of-the-art diffusion and auto-regressive models generally take much longer to train and generate, while GANs are much more efficient. The non-neural network model is fast to train but very slow to generate. The results verify the efficiency of GANs. 

\paragraph{Efficiency: Model Size.} When it comes to model sizes, the better-performing diffusion models and auto-regressive models typically involve a transformer, with 20M or 80M of parameters usually. In comparison, TAEGAN involves only networks of simple MLP structure and the total model size including the discriminator is less than 0.5M. This also showcases the outstanding efficiency of TAEGAN.

The detailed experimental results of all metrics can be found in Appendix~\ref{app:exp:baseline}.

\subsection{Loss Dynamics}

Fig.~\ref{fig:loss} shows the trend of losses on two datasets. The reconstruction loss hits a plateau during warmup and continues to decrease when the main training stage starts, validating the usefulness of GAN framework besides a standalone auto-encoder. The difference of discriminator output on real and synthetic data is generally stable, verifying the stability of TAEGAN training.

\subsection{Ablation Study}
\begin{wraptable}{r}{0.6\linewidth}
\vspace{-4.5em}
    \caption{Ablation study results. }
    \vspace{-1em}
    \label{tab:abl}
    \resizebox{\linewidth}{!}{
    \begin{tabular}{l|cccc|cc}
\toprule
 & \multicolumn{4}{c|}{MLE} & \multicolumn{2}{c}{Fidelity} \\
 & LG & RF & XGB & All & MD & CR \\
\midrule
TAEGAN & 0.867 & 0.861 & 0.866 & 0.865 & 0.952 & 0.933 \\
\midrule
w/o warmup & 0.861 & 0.848 & 0.857 & 0.855 & 0.954 & 0.928 \\
w/o log. freq. & 0.868 & 0.859 & 0.859 & 0.862 & 0.957 & 0.938 \\
$\lambda_1=\lambda_2=1$ & 0.859 & 0.858 & 0.852 & 0.857 & 0.953 & 0.933 \\
w/o discrete noise & 0.864 & 0.859 & 0.864 & 0.862 & 0.952 & 0.929 \\
w/o info. loss & 0.867 & 0.861 & 0.861 & 0.863 & 0.952 & 0.933 \\
\bottomrule
\end{tabular}

    }
    \vspace{-1em}
\end{wraptable}

In our ablation study, we assess the impact of key design choices in TAEGAN by systematically removing components. We highlight the result of the following 5 settings:
i) Warmup training is eliminated (\textbf{w/o warmup}) by reallocating warmup epochs to main training; 
ii) Data sampling during training is adjusted to {uniform sampling} (\textbf{w/o log. freq.}) rather than logarithmic frequency-based sampling;
iii) Mask ratio is fixed at equal values (\textbf{$\lambda_1 =$ $\lambda_2 =$ 1}); iv); Noise representation is altered by converting discrete noise to continuous values (\textbf{w/o discrete noise}); v) Information loss is removed (\textbf{w/o info. loss}).
Ablation experiments are carried out on 3 datasets: \texttt{adult}, \texttt{credit}, and \texttt{diabetes}.
TAEGAN with full setting performs the best on MLE, validating the effectiveness of all design components. The effect of warmup (vs. w/o warmup) and adjusted weights by mask ratio (vs. $\lambda_1=\lambda_2=1$) are particularly obvious. Fidelity scores are better without logarithmically transformed sampling (w/o log. freq.), which is consistent with our supposition in Section~\ref{sec:exp:exp} on the reason for the suboptimal performance of TAEGAN in fidelity.

\section{Conclusion}
\label{sec:conc}

In this paper, we introduce TAEGAN, a novel GAN framework for tabular data generation that uses a masked auto-encoder as the generator. We also propose a logarithmic frequency-based sampling method to address data imbalance and skewness, along with an improved loss function for better distribution and correlation in TAEGAN. When compared to existing baseline GANs and other tabular generative models across 8 datasets, synthetic data generated by TAEGAN achieves the highest utility in downstream tasks, closely resembles real data, while still effectively preserving privacy.



\bibliography{ref}

\clearpage
\appendix
\section{Details of TAEGAN}
\label{app:taegan}
\begin{algorithm}[H]
\small

\LinesNumbered
\SetAlgoNoEnd
            \captionsetup{type=algobox}
\captionof{algobox}{TAEGAN Generation Algorithm}
\label{alg:sample}
\KwData{Training table $\mathcal{T}$, generator parameters $\boldsymbol{\theta}_G$}
\KwResult{Generated row $\mathbf{x}'$}
$\boldsymbol{\sigma}\gets$ randomly generated order (permutation of $[1,2,\dots,C]$ with first $D_d$ being categorical indices)\;
$\mathbf{m}\gets\mathbf{0},\mathbf{m}_{\boldsymbol{\sigma}_1}\gets1$ \Comment*[r]{Initial mask with one selected only}
$\mathbf{x}'\gets$ randomly sampled $\boldsymbol{\gamma}$ with the current $\mathbf{m}$\;
\For{$i$ \textbf{in} $2,3,\dots,C$}{
    $\boldsymbol{\gamma}\gets\boldsymbol{\mu}^T\cdot\mathbf{x}'$ with $\boldsymbol{\mu}$ constructed based on $\mathbf{m}$\;
    Generate $\mathbf{x}'\gets G(\mathbf{m},\boldsymbol{\gamma},\mathbf{z})$ based on noise $\mathbf{z}$ from latent space\;
    $\mathbf{m}_{\boldsymbol{\sigma}_i}\gets1$ \Comment*[r]{Update mask with an additional component}
}
\end{algorithm}

\begin{figure}[t]
    \centering
    \begin{minipage}{0.49\linewidth}
        \centering
        \includegraphics[width=\linewidth]{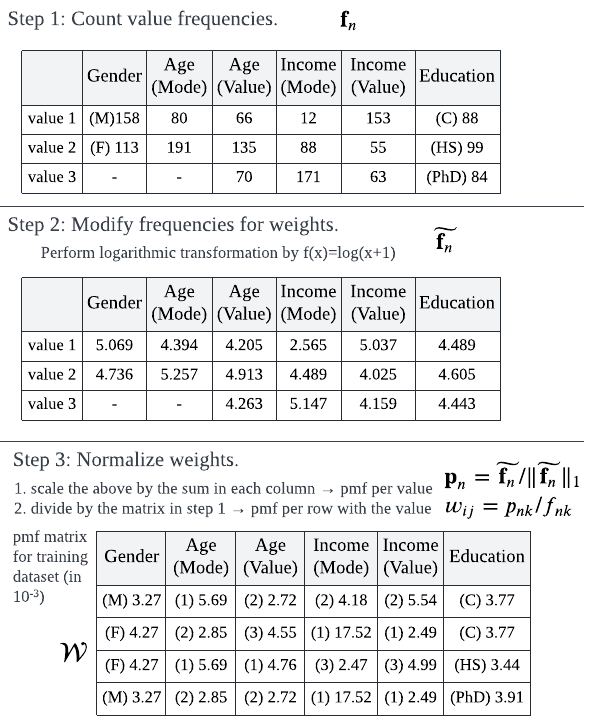}
        \caption{\textit{Weight matrix} calculation explained with the example.}
        \label{fig:weight-matrix}
    \end{minipage}
    \hfill
    \begin{minipage}{0.49\linewidth}
        \centering
        \includegraphics[width=\linewidth]{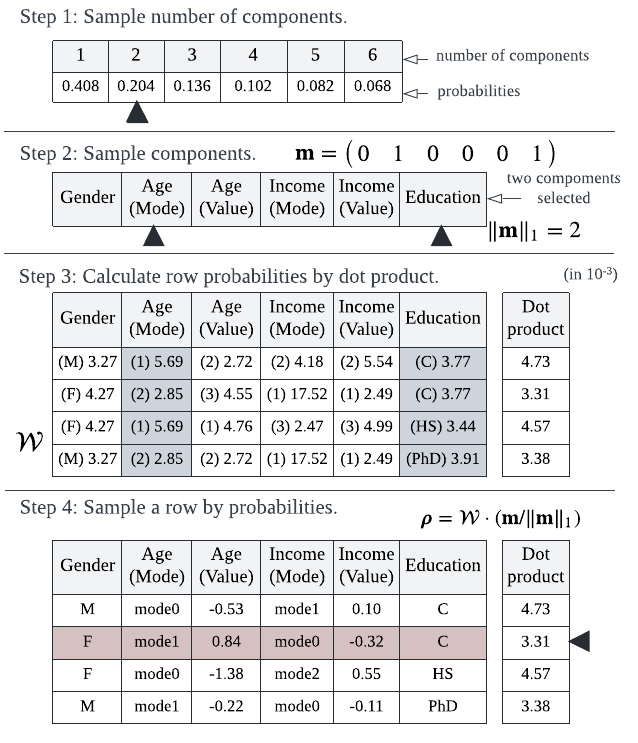}
        \caption{Multivariate data sampling using \textit{weight matrix} explained with the example.}
        \label{fig:new-sample}
    \end{minipage}
\end{figure}

\begin{wraptable}{r}{0.7\linewidth}
    \centering
    \vspace{-1.5em}
    \caption{Summary of the notations related to \textit{conditional vector}s.}
    \vspace{-1em}
    \label{tab:cond-dim}
    \resizebox{\linewidth}{!}{
    \begin{tabular}{L{0.55\linewidth}|l|L{0.22\linewidth}|L{0.2\linewidth}}
    \toprule
    component $\alpha$ & $\mathbf{m}_{\alpha}$ & $\boldsymbol{\mu}_{\alpha}$ & $\boldsymbol{\gamma}_{\alpha}$\\
    \midrule
    maintained after mask & $(1)$ & $\mathbf{1}\in\{0,1\}^{D_{\alpha}}$ & $\mathbf{x}_{\alpha}$ ($\mathbf{c}_n$ or $\mathbf{d}_n$) \\\hline
    not maintained after mask & $(0)$ & $\mathbf{0}\in\{0,1\}^{D_{\alpha}}$ & $\mathbf{0}\in\mathbb{R}^{D_{\alpha}}$ \\\hline
    empty / disallowed in condition & $\boldsymbol{\phi}$ & $\boldsymbol{\phi}$ & $\boldsymbol{\phi}$ \\
    \bottomrule
\end{tabular}
    }
    \vspace{-1em}
\end{wraptable}

The notations related to the construction of \textit{conditional vectors} introduced in Section~\ref{sec:pre:cond} are summarized in Table~\ref{tab:cond-dim}.

The TAEGAN generation process described in Section~\ref{sec:taegan:overview} and visualized in Fig.~\ref{fig:gen} can be formally described as the pseudocode in Algorithm~\ref{alg:sample}. 

Fig.~\ref{fig:weight-matrix}-\ref{fig:new-sample} provides an illustrative example with actual values of \textit{weight matrix} calculation (Algorithm~\ref{alg:weight}) and data sampling based on it (Algorithm~\ref{alg:log-sample}) described in Appendix~\ref{sec:taegan:sample}. This Fig. aims to aid understanding of the process of data sampling based on logarithmic frequency during training.

\section{Proof of Proposition~\ref{thm:log-ctgan}}
\label{app:proof}
\begin{proof}
    In CTGAN, $\mathbb{P}(\boldsymbol{\gamma}|\mathbf{m})$ is derived from $\mathbf{p}$ of Algorithm~\ref{alg:weight} (recall Section~\ref{sec:pre:sample}), based on the value in the selected component by $\mathbf{m}$.

    In TAEGAN, 
    \begin{align}
        &\mathbb{P}(\boldsymbol{\gamma}|\mathbf{m})\notag\\
        = &\mathbb{P}(\boldsymbol{\gamma}=\boldsymbol{\mu}^T\cdot\mathbf{x}|\mathbf{m})\notag\\
        = &\mathbb{P}(\text{a row with }\boldsymbol{\mu}^T\cdot\mathbf{x}=\gamma\text{ is sampled})\cdot(\text{number of rows with }\boldsymbol{\mu}^T\cdot\mathbf{x}=\gamma)\notag\\
        = &\widetilde{\mathbf{p}}^T\cdot\mathbf{f} \text{ from Algorithm~\ref{alg:weight} mapped to each row's value}\notag\\
        = &\mathbf{p} \text{ from Algorithm~\ref{alg:weight} mapped to each row's value}
    \end{align}
    Therefore, these two $\mathbb{P}(\boldsymbol{\gamma}|\mathbf{m})$ values are identical.
\end{proof}

\section{Experiment Setup}
\subsection{Datasets}
\label{app:setup:datasets}
Table~\ref{tab:datasets} summarizes the 8 datasets used. All datasets are classification datasets for easiness of benchmarking. Datasets are downloaded by \verb|sklearn.datasets.fetch_openml| with the dataset name as input.

\subsection{Baselines Implementation}
\label{app:setup:baselines}


\begin{table}[htbp]
    \centering
    \caption{Datasets used in experiments. \#R, \#F, \#N, \#D, and \#C represent the number of rows, 
    features (including target column), numeric features, discrete (i.e., categorical) features, 
    and classes respectively. Aliases in brackets show the names used in this paper.}
    \label{tab:datasets}

    \resizebox{0.7\textwidth}{!}{
        \begin{tabular}{lrrrrr}
\toprule
Name (Alias) & \#R & \#F & \#N & \#D & \#C \\
\midrule
adult & 48842 & 15 & 2 & 13 & 2 \\
bank-marketing (bank) & 45211 & 17 & 7 & 10 & 2 \\
breast-w (breast) & 699 & 10 & 9 & 1 & 2 \\
credit-g (credit) & 1000 & 21 & 7 & 14 & 2 \\
diabetes & 768 & 9 & 8 & 1 & 2 \\
iris & 150 & 5 & 4 & 1 & 3 \\
qsar-biodeg (qsar) & 1055 & 42 & 41 & 1 & 2 \\
wdbc & 569 & 31 & 30 & 1 & 2 \\
\bottomrule
\end{tabular}

    }
\end{table}

We use the Synthcity~\citep{synthcity} implementation for ARF~\citep{arf}, CTGAN~\citep{ctgan}, GReaT~\citep{great}, TabDDPM~\citep{tabddpm}, and TVAE~\citep{ctgan}. We also include CTAB-GAN+~\citep{ctabganp} and TabSyn~\citep{tabsyn} using their official code on GitHub. 

\subsection{TAEGAN Configuration}
\label{app:setup:taegan}

\textbf{Model dimensions and layers.} The dimension of $\mathbf{z}$ is set to 100, among which 50 are discrete and 50 are continuous. The encoder (of the generator), decoder (of the generator), and discriminator are all MLPs with 6 layers and a hidden dimension of 128. Each layer consists of a batch normalization~\citep{batch-norm} and is activated by ReLU except for the last layer. The encoded dimension of the generator (i.e., output of encoder) is 256. All MLPs are trained with a dropout rate of 0.1.

\textbf{Training.} In each step, the discriminator is updated three times instead of once as shown in Algorithm~\ref{alg:train}, which is aimed for higher-level illustration. Updating the discriminator more frequently than generator is a common practice for GANs trained with WGAN-GP~\citep{wgan-gp}. Usually, the discriminator is updated five times in each step for WGAN-GP, but we have a stronger warmed up masked auto-encoder as the generator, so we reduces this value. During training, we apply a weight decay of $1\times10^{-5}$ and gradient norm of 5.

\textbf{Other hyperparameters.} We set $\lambda_1=0.1,\lambda_2=1,\alpha_1=1$. The \textit{outer product} is symmetric, and we use it for correlation, so it suffices to use the upper triangular only for the loss calculation. Both the mean loss and interaction loss's weight are the inverse of their dimensions (e.g., $\alpha_2=\frac{1}{C}$).

\subsection{MLE Details}
\label{app:setup:mle}
Hyperparameters of downstream tasks using random forest and XGBoost are tuned by Optuna~\citep{optuna} in 30 trials.

The hyperparameter space explored for random forest is
\begin{itemize}
    \item \textbf{Number of Estimators}: Integers in $[100,300]$ step 50;
    \item \textbf{Maximum Depth}: Integers in $[5, 20]$ step 5;
    \item \textbf{Minimum Samples Split}: Integers in $[2,10]$ step 2;
    \item \textbf{Minimum Size per Leaf}: Integers $[1,5]$;
    \item \textbf{Maximum Features}: Value in ``sqrt'', ``log2'', and NULL;
    \item \textbf{Bootstramp}: Enabled or disabled.
\end{itemize}

The hyperparameter space explored for XGBoost is:
\begin{itemize}
    \item \textbf{Learning Rate}: Logarithmic scale float in $[0.01,0.3]$;
    \item \textbf{Number of Estimators}: Integers in $[100,300]$ step 50;
    \item \textbf{Maximum Depth}: Integers in $[3,10]$;
    \item \textbf{Minimum Child Weight}: Logarithmic scale float in $[1.0,10.0]$;
    \item \textbf{Minimum Split Loss Gamma}: Float in $[0.0,0.5]$;
    \item \textbf{Subsample Ratio}: Float in $[0.5,1.0]$;
    \item \textbf{Subsample Ratio of Columns by Tree}: Float in $[0.5,1.0]$;
    \item \textbf{L1 Regularization}: Float in $[0.0,10.0]$.
\end{itemize}

For all models, numeric values are standardized using standard scaling. Categorical values are one-hot encoded for logistic regression and label-encoded for random forest and XGBoost. To simplify the experiments, rows with missing values are excluded, as handling missing data is not the focus of this paper.

\subsection{RSD Details}
The training data for the RSD task consists of the real training set combined with a synthetic dataset of the same size as the real training set. The test data comprises the real test set combined with another synthetic dataset of the same size as the real test set. The implementation of ML tasks follow from the MLE implementation.

\subsection{Fidelity Details}
Fidelity metrics are calculated using SDMetrics~\citep{sdmetrics}'s public SDK, where MD is evaluated by ``Shape'' and DC is evaluated by ``Trend'' scores.

\subsection{DCR Details}

Let the real training dataset be $\mathbf{X}$, real hold-out test set be $\widehat{\mathbf{X}}$, and a synthetic dataset of the same size be $\mathbf{X}'$.
We compare the DCR of $(\mathbf{X},\widehat{\mathbf{X}})$ and $(\mathbf{X},\mathbf{X}')$. The synthetic data is deemed privacy-preserving if the former values are not smaller than the latter. This is tested using the Mann-Whitney U Test~\citep{mwu}, where the null hypothesis \(H_0\) states that the distance between \(\widehat{\mathbf{X}}\) and \(\mathbf{X}\) is greater than or equal to the distance between \(\mathbf{X}'\) and \(\mathbf{X}\). If the \(p\)-value is below 0.05, it indicates that \(\mathbf{X}'\) is closer to \(\mathbf{X}\) than \(\widehat{\mathbf{X}}\), suggesting a potential risk of privacy leakage.

For distance calculations, the data is preprocessed by standard scaling numeric features and applying one-hot encoding to categorical features. Cosine distance is employed to measure the distance between records. As a baseline, we calculate the cosine distances between the $\widehat{\mathbf{X}}$ and the $\mathbf{X}$, recording the minimum distance for each record in $\widehat{\mathbf{X}}$. Similarly, we compute the cosine distances between the $\mathbf{X}'$ and $\mathbf{X}$, also extracting the minimum distances. These distances are then compared to the baseline values to evaluate the similarity and privacy-preserving properties of the synthetic data.

\begin{table}[t]
    \centering
    \caption{Ablation experiments setup.}
    \label{tab:abl-setup}
    \begin{tabular}{lp{0.8\linewidth}}
    \toprule
    Short form & Description \\
    \midrule
    w/o warmup & No warmup, warmup epochs are changed to main training epochs \\
    gen. in 2 steps & Generating data in 2 steps instead of one component per step ($C$ steps) \\
    w/o log. freq. & Data sampling during training by uniform sampling per row instead of using logarithmic frequency \\
    $\lambda_1=\lambda_2=1$ & Uniform weight for different mask ratio\\
    w/o discrete noise & Discrete noise dimensions are set to continuous \\
    w/o pac & Training without PacGAN framework, i.e., pac size is 1\\
    w/o info. loss & Weights of all components of information loss set to 0 \\
    w/o interac. loss & Weights of interaction loss set to 0\\
    \bottomrule
\end{tabular}
\end{table}

\subsection{Efficiency Details}
\label{app:setup:efficiency}

The time records for training and generation are computed from the recorded start and end times of the operations.

\subsection{Ablation Experiment Setup}
\label{app:setup:abl}
For experiment efficiency, we run ablation experiments on only 3 datasets: \texttt{adult}, \texttt{credit}, and \texttt{diabetes}. The ablation experiments' configurations are shown in Table~\ref{tab:abl-setup}. More settings are provided than the main paper.

\section{Detailed Experiment Results}
\label{app:exp}

\subsection{Baseline Comparison}
\label{app:exp:baseline}

\begin{table}[t]
    \centering
    \caption{Raw MLE scores in each dataset. All values indicate better performance when the score is higher, except for RE (relative error), which is better when the score is lower. The best scores are highlighted in bold and underscore, and the second best scores are highlighted in bold.}
    \label{tab:mle-details}
    \resizebox{\linewidth}{!}{
    \begin{tabular}{ll|c|ccccccc|c}
\toprule
Dataset & ML & real & ARF & CTAB+ & CTGAN & TDDPM & GReaT & TabSyn & TVAE & TAEGAN \\
\midrule
\multirow{3}{*}{adult} & LG & 0.914 & $0.907_{\pm 0.000}$ & $0.904_{\pm 0.001}$ & $0.886_{\pm 0.009}$ & $0.908_{\pm 0.000}$ & $\underline{\boldsymbol{0.911}}_{\pm 0.000}$ & $\boldsymbol{0.910}_{\pm 0.002}$ & $0.893_{\pm 0.001}$ & $0.907_{\pm 0.001}$ \\
 & RF & 0.910 & $0.903_{\pm 0.001}$ & $0.902_{\pm 0.002}$ & $0.884_{\pm 0.003}$ & $0.904_{\pm 0.001}$ & $\underline{\boldsymbol{0.905}}_{\pm 0.000}$ & $\boldsymbol{0.905}_{\pm 0.003}$ & $0.892_{\pm 0.001}$ & $0.905_{\pm 0.002}$ \\
 & XGB & 0.914 & $0.907_{\pm 0.001}$ & $0.905_{\pm 0.001}$ & $0.886_{\pm 0.002}$ & $0.908_{\pm 0.000}$ & $\underline{\boldsymbol{0.911}}_{\pm 0.000}$ & $\boldsymbol{0.910}_{\pm 0.003}$ & $0.895_{\pm 0.001}$ & $0.909_{\pm 0.001}$ \\
\cline{1-11}
\multirow{3}{*}{bank} & LG & 0.907 & $0.895_{\pm 0.002}$ & $\boldsymbol{0.901}_{\pm 0.003}$ & $0.887_{\pm 0.003}$ & $0.900_{\pm 0.002}$ & $\underline{\boldsymbol{0.907}}_{\pm 0.000}$ & $0.900_{\pm 0.005}$ & $0.890_{\pm 0.005}$ & $0.899_{\pm 0.000}$ \\
 & RF & 0.930 & $0.900_{\pm 0.001}$ & $0.902_{\pm 0.002}$ & $0.882_{\pm 0.002}$ & $\boldsymbol{0.908}_{\pm 0.001}$ & $0.907_{\pm 0.001}$ & $0.907_{\pm 0.006}$ & $0.884_{\pm 0.002}$ & $\underline{\boldsymbol{0.914}}_{\pm 0.001}$ \\
 & XGB & 0.936 & $0.903_{\pm 0.005}$ & $0.906_{\pm 0.000}$ & $0.884_{\pm 0.002}$ & $0.912_{\pm 0.001}$ & $0.910_{\pm 0.001}$ & $\boldsymbol{0.913}_{\pm 0.007}$ & $0.885_{\pm 0.001}$ & $\underline{\boldsymbol{0.920}}_{\pm 0.001}$ \\
\cline{1-11}
\multirow{3}{*}{breast} & LG & 0.985 & $0.984_{\pm 0.005}$ & $0.913_{\pm 0.091}$ & $0.978_{\pm 0.006}$ & $0.985_{\pm 0.002}$ & $0.985_{\pm 0.000}$ & $\boldsymbol{0.987}_{\pm 0.001}$ & $0.930_{\pm 0.009}$ & $\underline{\boldsymbol{0.990}}_{\pm 0.001}$ \\
 & RF & 0.985 & $0.982_{\pm 0.003}$ & $0.968_{\pm 0.012}$ & $\boldsymbol{0.985}_{\pm 0.003}$ & $0.981_{\pm 0.002}$ & $0.979_{\pm 0.004}$ & $0.981_{\pm 0.002}$ & $0.976_{\pm 0.007}$ & $\underline{\boldsymbol{0.985}}_{\pm 0.003}$ \\
 & XGB & 0.984 & $0.979_{\pm 0.001}$ & $0.958_{\pm 0.025}$ & $\boldsymbol{0.984}_{\pm 0.002}$ & $0.981_{\pm 0.001}$ & $0.982_{\pm 0.002}$ & $\underline{\boldsymbol{0.987}}_{\pm 0.001}$ & $0.981_{\pm 0.005}$ & $0.983_{\pm 0.001}$ \\
\cline{1-11}
\multirow{3}{*}{credit} & LG & 0.836 & $0.773_{\pm 0.064}$ & $0.538_{\pm 0.093}$ & $\boldsymbol{0.814}_{\pm 0.012}$ & $0.781_{\pm 0.018}$ & $0.665_{\pm 0.000}$ & $0.780_{\pm 0.017}$ & $0.702_{\pm 0.042}$ & $\underline{\boldsymbol{0.820}}_{\pm 0.015}$ \\
 & RF & 0.837 & $0.769_{\pm 0.030}$ & $0.505_{\pm 0.097}$ & $0.782_{\pm 0.033}$ & $0.755_{\pm 0.013}$ & $0.722_{\pm 0.014}$ & $\boldsymbol{0.791}_{\pm 0.007}$ & $0.746_{\pm 0.057}$ & $\underline{\boldsymbol{0.819}}_{\pm 0.025}$ \\
 & XGB & 0.844 & $0.788_{\pm 0.019}$ & $0.509_{\pm 0.088}$ & $0.788_{\pm 0.023}$ & $0.748_{\pm 0.026}$ & $0.739_{\pm 0.006}$ & $\boldsymbol{0.795}_{\pm 0.014}$ & $0.759_{\pm 0.048}$ & $\underline{\boldsymbol{0.833}}_{\pm 0.014}$ \\
\cline{1-11}
\multirow{3}{*}{diabetes} & LG & 0.884 & $0.854_{\pm 0.014}$ & $0.817_{\pm 0.028}$ & $0.864_{\pm 0.017}$ & $0.753_{\pm 0.032}$ & $0.858_{\pm 0.000}$ & $\underline{\boldsymbol{0.883}}_{\pm 0.006}$ & $0.863_{\pm 0.008}$ & $\boldsymbol{0.874}_{\pm 0.010}$ \\
 & RF & 0.869 & $0.818_{\pm 0.023}$ & $0.802_{\pm 0.014}$ & $0.818_{\pm 0.008}$ & $0.838_{\pm 0.024}$ & $0.824_{\pm 0.012}$ & $\boldsymbol{0.844}_{\pm 0.012}$ & $0.817_{\pm 0.020}$ & $\underline{\boldsymbol{0.860}}_{\pm 0.021}$ \\
 & XGB & 0.867 & $0.838_{\pm 0.012}$ & $0.811_{\pm 0.005}$ & $0.807_{\pm 0.005}$ & $0.832_{\pm 0.017}$ & $0.831_{\pm 0.003}$ & $\boldsymbol{0.847}_{\pm 0.009}$ & $0.820_{\pm 0.020}$ & $\underline{\boldsymbol{0.855}}_{\pm 0.010}$ \\
\cline{1-11}
\multirow{3}{*}{iris} & LG & 1.000 & $\boldsymbol{0.997}_{\pm 0.005}$ & $0.272_{\pm 0.068}$ & $0.886_{\pm 0.025}$ & $0.983_{\pm 0.015}$ & $0.985_{\pm 0.000}$ & $\underline{\boldsymbol{0.999}}_{\pm 0.002}$ & $0.937_{\pm 0.030}$ & $0.971_{\pm 0.015}$ \\
 & RF & 1.000 & $0.998_{\pm 0.002}$ & $0.396_{\pm 0.099}$ & $0.936_{\pm 0.018}$ & $0.934_{\pm 0.058}$ & $\underline{\boldsymbol{1.000}}_{\pm 0.000}$ & $\underline{\boldsymbol{1.000}}_{\pm 0.000}$ & $0.995_{\pm 0.004}$ & $0.997_{\pm 0.003}$ \\
 & XGB & 1.000 & $\underline{\boldsymbol{1.000}}_{\pm 0.000}$ & $0.211_{\pm 0.042}$ & $0.971_{\pm 0.009}$ & $0.817_{\pm 0.203}$ & $\underline{\boldsymbol{1.000}}_{\pm 0.000}$ & $\underline{\boldsymbol{1.000}}_{\pm 0.000}$ & $0.986_{\pm 0.004}$ & $0.997_{\pm 0.004}$ \\
\cline{1-11}
\multirow{3}{*}{qsar} & LG & 0.906 & $\underline{\boldsymbol{0.888}}_{\pm 0.011}$ & $0.716_{\pm 0.085}$ & $\boldsymbol{0.869}_{\pm 0.019}$ & $0.783_{\pm 0.057}$ & $0.673_{\pm 0.000}$ & $0.867_{\pm 0.012}$ & $0.862_{\pm 0.015}$ & $0.864_{\pm 0.016}$ \\
 & RF & 0.936 & $0.877_{\pm 0.021}$ & $0.648_{\pm 0.036}$ & $0.874_{\pm 0.020}$ & $0.714_{\pm 0.080}$ & $0.616_{\pm 0.009}$ & $\underline{\boldsymbol{0.882}}_{\pm 0.006}$ & $0.850_{\pm 0.026}$ & $\boldsymbol{0.882}_{\pm 0.008}$ \\
 & XGB & 0.921 & $\boldsymbol{0.872}_{\pm 0.011}$ & $0.731_{\pm 0.020}$ & $\underline{\boldsymbol{0.874}}_{\pm 0.010}$ & $0.713_{\pm 0.056}$ & $0.602_{\pm 0.020}$ & $0.860_{\pm 0.016}$ & $0.856_{\pm 0.022}$ & $0.869_{\pm 0.010}$ \\
\cline{1-11}
\multirow{3}{*}{wdbc} & LG & 0.993 & $0.979_{\pm 0.006}$ & $0.929_{\pm 0.053}$ & $0.983_{\pm 0.007}$ & $0.946_{\pm 0.028}$ & $0.979_{\pm 0.000}$ & $\underline{\boldsymbol{0.992}}_{\pm 0.001}$ & $0.979_{\pm 0.013}$ & $\boldsymbol{0.989}_{\pm 0.002}$ \\
 & RF & 0.976 & $0.972_{\pm 0.002}$ & $0.919_{\pm 0.024}$ & $0.974_{\pm 0.004}$ & $0.894_{\pm 0.051}$ & $0.976_{\pm 0.003}$ & $\boldsymbol{0.984}_{\pm 0.004}$ & $0.976_{\pm 0.001}$ & $\underline{\boldsymbol{0.984}}_{\pm 0.002}$ \\
 & XGB & 0.990 & $0.979_{\pm 0.001}$ & $0.930_{\pm 0.021}$ & $0.974_{\pm 0.008}$ & $0.512_{\pm 0.082}$ & $0.973_{\pm 0.003}$ & $\boldsymbol{0.986}_{\pm 0.002}$ & $0.977_{\pm 0.006}$ & $\underline{\boldsymbol{0.987}}_{\pm 0.006}$ \\
\midrule
\multirow{5}{*}{Avg.} & All & 0.930 & 0.907 & 0.750 & 0.895 & 0.850 & 0.868 & \textbf{0.913} & 0.890 & \textbf{\underline{0.917}} \\
                      & RE ($\downarrow$) & - & 2.632\% & 19.217\% & 3.873\% & 8.604\% & 6.897\% & \textbf{1.951\%} & 4.511\% & \textbf{\underline{1.422\%}}\\
                      & LG & 0.928 & 0.909 & 0.749 & 0.896 & 0.880 & 0.870 & \textbf{\underline{0.915}} & 0.882 & \textbf{0.914} \\
                      & RF & 0.930 & 0.903 & 0.755 & 0.892 & 0.866 & 0.866 & \textbf{0.912} & 0.892 & \textbf{\underline{0.918}} \\
                      & XGB & 0.932 & 0.908 & 0.745 & 0.896 & 0.803 & 0.869 & \textbf{0.912} & 0.895 & \textbf{\underline{0.919}} \\
                      \bottomrule
\end{tabular}

    }
\end{table}

Table~\ref{tab:mle-details} shows the raw scores of MLE comparing different baseline models with TAEGAN. TAEGAN is better than all baseline models in general. In particular, TAEGAN is significantly better than baseline GAN models.

\begin{table}[t]
    \centering
    \caption{Raw RSD scores in each dataset. All values indicate better performance when the score is lower (capped at 0.5). The best scores are highlighted in bold and underscore, and the second best scores are highlighted in bold.}
    \label{tab:rsd-details}
    \resizebox{\linewidth}{!}{
    \begin{tabular}{ll|ccccccc|c}
\toprule
Dataset & ML & ARF & CTAB+ & CTGAN & TDDPM & GReaT & TabSyn & TVAE & TAEGAN \\
\midrule
\multirow{3}{*}{adult} & LG & $0.794_{\pm 0.006}$ & $0.632_{\pm 0.016}$ & $0.946_{\pm 0.005}$ & $\underline{\boldsymbol{0.585}}_{\pm 0.005}$ & $0.714_{\pm 0.000}$ & $\boldsymbol{0.614}_{\pm 0.031}$ & $0.945_{\pm 0.003}$ & $0.619_{\pm 0.005}$ \\
 & RF & $0.877_{\pm 0.003}$ & $0.711_{\pm 0.010}$ & $0.983_{\pm 0.007}$ & $0.739_{\pm 0.002}$ & $0.741_{\pm 0.001}$ & $\boldsymbol{0.691}_{\pm 0.067}$ & $0.978_{\pm 0.005}$ & $\underline{\boldsymbol{0.670}}_{\pm 0.001}$ \\
 & XGB & $0.902_{\pm 0.003}$ & $0.734_{\pm 0.011}$ & $0.989_{\pm 0.004}$ & $0.762_{\pm 0.002}$ & $0.758_{\pm 0.001}$ & $\boldsymbol{0.714}_{\pm 0.068}$ & $0.985_{\pm 0.003}$ & $\underline{\boldsymbol{0.695}}_{\pm 0.003}$ \\
\cline{1-10}
\multirow{3}{*}{bank} & LG & $0.736_{\pm 0.004}$ & $0.607_{\pm 0.013}$ & $0.818_{\pm 0.013}$ & $\boldsymbol{0.581}_{\pm 0.004}$ & $0.700_{\pm 0.000}$ & $0.588_{\pm 0.025}$ & $0.826_{\pm 0.021}$ & $\underline{\boldsymbol{0.573}}_{\pm 0.006}$ \\
 & RF & $0.869_{\pm 0.014}$ & $0.745_{\pm 0.007}$ & $0.961_{\pm 0.006}$ & $0.783_{\pm 0.003}$ & $0.824_{\pm 0.000}$ & $\boldsymbol{0.742}_{\pm 0.060}$ & $0.968_{\pm 0.006}$ & $\underline{\boldsymbol{0.657}}_{\pm 0.002}$ \\
 & XGB & $0.901_{\pm 0.008}$ & $0.795_{\pm 0.008}$ & $0.980_{\pm 0.002}$ & $0.821_{\pm 0.007}$ & $0.842_{\pm 0.001}$ & $\boldsymbol{0.784}_{\pm 0.038}$ & $0.983_{\pm 0.002}$ & $\underline{\boldsymbol{0.703}}_{\pm 0.004}$ \\
\cline{1-10}
\multirow{3}{*}{breast} & LG & $\underline{\boldsymbol{0.605}}_{\pm 0.006}$ & $0.723_{\pm 0.014}$ & $0.690_{\pm 0.040}$ & $\boldsymbol{0.619}_{\pm 0.013}$ & $0.685_{\pm 0.000}$ & $0.695_{\pm 0.032}$ & $0.702_{\pm 0.046}$ & $0.682_{\pm 0.034}$ \\
 & RF & $\underline{\boldsymbol{0.615}}_{\pm 0.014}$ & $0.921_{\pm 0.041}$ & $0.743_{\pm 0.021}$ & $\boldsymbol{0.623}_{\pm 0.006}$ & $0.668_{\pm 0.001}$ & $0.670_{\pm 0.019}$ & $0.823_{\pm 0.012}$ & $0.661_{\pm 0.023}$ \\
 & XGB & $\underline{\boldsymbol{0.604}}_{\pm 0.011}$ & $0.922_{\pm 0.041}$ & $0.751_{\pm 0.026}$ & $\boldsymbol{0.611}_{\pm 0.005}$ & $0.652_{\pm 0.017}$ & $0.700_{\pm 0.018}$ & $0.850_{\pm 0.017}$ & $0.674_{\pm 0.039}$ \\
\cline{1-10}
\multirow{3}{*}{credit} & LG & $\underline{\boldsymbol{0.505}}_{\pm 0.007}$ & $0.634_{\pm 0.066}$ & $0.654_{\pm 0.035}$ & $0.663_{\pm 0.013}$ & $0.708_{\pm 0.000}$ & $0.812_{\pm 0.034}$ & $0.688_{\pm 0.027}$ & $\boldsymbol{0.564}_{\pm 0.009}$ \\
 & RF & $0.858_{\pm 0.003}$ & $0.910_{\pm 0.074}$ & $0.917_{\pm 0.017}$ & $0.794_{\pm 0.026}$ & $0.816_{\pm 0.001}$ & $\boldsymbol{0.785}_{\pm 0.038}$ & $0.921_{\pm 0.011}$ & $\underline{\boldsymbol{0.608}}_{\pm 0.028}$ \\
 & XGB & $0.866_{\pm 0.004}$ & $0.920_{\pm 0.083}$ & $0.927_{\pm 0.013}$ & $\boldsymbol{0.799}_{\pm 0.021}$ & $0.835_{\pm 0.005}$ & $0.856_{\pm 0.017}$ & $0.942_{\pm 0.008}$ & $\underline{\boldsymbol{0.640}}_{\pm 0.032}$ \\
\cline{1-10}
\multirow{3}{*}{diabetes} & LG & $0.547_{\pm 0.009}$ & $0.629_{\pm 0.045}$ & $0.767_{\pm 0.026}$ & $0.714_{\pm 0.012}$ & $0.727_{\pm 0.000}$ & $\underline{\boldsymbol{0.469}}_{\pm 0.021}$ & $0.573_{\pm 0.018}$ & $\boldsymbol{0.524}_{\pm 0.029}$ \\
 & RF & $0.725_{\pm 0.021}$ & $0.845_{\pm 0.016}$ & $0.924_{\pm 0.038}$ & $0.865_{\pm 0.017}$ & $0.811_{\pm 0.001}$ & $\underline{\boldsymbol{0.603}}_{\pm 0.022}$ & $0.953_{\pm 0.009}$ & $\boldsymbol{0.677}_{\pm 0.009}$ \\
 & XGB & $0.754_{\pm 0.018}$ & $0.844_{\pm 0.014}$ & $0.948_{\pm 0.019}$ & $0.868_{\pm 0.011}$ & $0.814_{\pm 0.003}$ & $\underline{\boldsymbol{0.706}}_{\pm 0.012}$ & $0.977_{\pm 0.009}$ & $\boldsymbol{0.707}_{\pm 0.022}$ \\
\cline{1-10}
\multirow{3}{*}{iris} & LG & $0.518_{\pm 0.077}$ & $0.796_{\pm 0.040}$ & $0.867_{\pm 0.002}$ & $0.806_{\pm 0.111}$ & $0.534_{\pm 0.000}$ & $\underline{\boldsymbol{0.485}}_{\pm 0.081}$ & $\boldsymbol{0.510}_{\pm 0.055}$ & $0.541_{\pm 0.096}$ \\
 & RF & $0.877_{\pm 0.023}$ & $0.989_{\pm 0.008}$ & $0.969_{\pm 0.007}$ & $0.999_{\pm 0.002}$ & $\underline{\boldsymbol{0.628}}_{\pm 0.029}$ & $0.896_{\pm 0.067}$ & $0.954_{\pm 0.014}$ & $\boldsymbol{0.665}_{\pm 0.024}$ \\
 & XGB & $0.931_{\pm 0.037}$ & $0.984_{\pm 0.010}$ & $0.944_{\pm 0.003}$ & $0.998_{\pm 0.003}$ & $\underline{\boldsymbol{0.570}}_{\pm 0.009}$ & $0.896_{\pm 0.052}$ & $0.975_{\pm 0.006}$ & $\boldsymbol{0.871}_{\pm 0.015}$ \\
\cline{1-10}
\multirow{3}{*}{qsar} & LG & $\underline{\boldsymbol{0.659}}_{\pm 0.025}$ & $0.838_{\pm 0.031}$ & $0.930_{\pm 0.013}$ & $1.000_{\pm 0.000}$ & $0.821_{\pm 0.000}$ & $0.770_{\pm 0.016}$ & $0.801_{\pm 0.034}$ & $\boldsymbol{0.693}_{\pm 0.002}$ \\
 & RF & $0.972_{\pm 0.006}$ & $1.000_{\pm 0.000}$ & $0.999_{\pm 0.001}$ & $1.000_{\pm 0.000}$ & $0.979_{\pm 0.001}$ & $\boldsymbol{0.946}_{\pm 0.018}$ & $0.998_{\pm 0.001}$ & $\underline{\boldsymbol{0.838}}_{\pm 0.019}$ \\
 & XGB & $0.984_{\pm 0.007}$ & $1.000_{\pm 0.000}$ & $0.999_{\pm 0.000}$ & $1.000_{\pm 0.000}$ & $0.989_{\pm 0.002}$ & $\boldsymbol{0.975}_{\pm 0.003}$ & $0.998_{\pm 0.001}$ & $\underline{\boldsymbol{0.869}}_{\pm 0.009}$ \\
\cline{1-10}
\multirow{3}{*}{wdbc} & LG & $\boldsymbol{0.517}_{\pm 0.008}$ & $0.832_{\pm 0.049}$ & $0.961_{\pm 0.020}$ & $0.994_{\pm 0.004}$ & $0.675_{\pm 0.000}$ & $0.658_{\pm 0.018}$ & $0.743_{\pm 0.024}$ & $\underline{\boldsymbol{0.510}}_{\pm 0.037}$ \\
 & RF & $0.948_{\pm 0.007}$ & $1.000_{\pm 0.000}$ & $0.993_{\pm 0.006}$ & $1.000_{\pm 0.000}$ & $0.965_{\pm 0.003}$ & $\underline{\boldsymbol{0.616}}_{\pm 0.057}$ & $0.989_{\pm 0.001}$ & $\boldsymbol{0.633}_{\pm 0.050}$ \\
 & XGB & $0.939_{\pm 0.006}$ & $0.999_{\pm 0.000}$ & $0.990_{\pm 0.008}$ & $1.000_{\pm 0.000}$ & $0.968_{\pm 0.004}$ & $\underline{\boldsymbol{0.677}}_{\pm 0.065}$ & $0.994_{\pm 0.001}$ & $\boldsymbol{0.698}_{\pm 0.018}$ \\
\midrule
\multirow{4}{*}{Avg.} & All & 0.771 & 0.834 & 0.902 & 0.818 & 0.768 & \textbf{0.723} & 0.878 & \textbf{\underline{0.666}} \\
                      & LG & \textbf{0.610} & 0.711 & 0.829 & 0.745 & 0.696 & 0.636 & 0.724 & \textbf{\underline{0.588}} \\
                      & RF & 0.843 & 0.890 & 0.936 & 0.850 & 0.804 & \textbf{0.743} & 0.948 & \textbf{\underline{0.676}} \\
                      & XGB & 0.860 & 0.900 & 0.941 & 0.857 & 0.803 & \textbf{0.789} & 0.963 & \textbf{\underline{0.732}} \\
                      \bottomrule
\end{tabular}

    }
\end{table}

Table~\ref{tab:rsd-details} shows the raw scores of RSD comparing different baseline models with TAEGAN. TAEGAN demonstrates wins over all baseline models by a great margin.

\begin{table}[t]
    \centering
    \caption{Raw fidelity scores in each dataset. All values indicate better performance when the score is higher. The best scores are highlighted in bold and underscore, and the second best scores are highlighted in bold. The best scores among GANs are highlighted with superscript ${}^*$.}
    \label{tab:fidelity-details}
    \resizebox{\linewidth}{!}{
    \begin{tabular}{ll|ccccccc|c}
\toprule
Dataset & Metric & ARF & CTAB+ & CTGAN & TDDPM & GReaT & TabSyn & TVAE & TAEGAN \\
\midrule
\multirow{2}{*}{adult} & MD & $0.942_{\pm 0.001}$ & $0.964_{\pm 0.005}^*$ & $0.881_{\pm 0.008}$ & $\underline{\boldsymbol{0.983}}_{\pm 0.001}$ & $0.929_{\pm 0.000}$ & $\boldsymbol{0.977}_{\pm 0.011}$ & $0.877_{\pm 0.004}$ & $0.958_{\pm 0.002}$ \\
 & CR & $0.876_{\pm 0.004}$ & $0.915_{\pm 0.004}$ & $0.760_{\pm 0.005}$ & $\boldsymbol{0.953}_{\pm 0.006}$ & $0.882_{\pm 0.000}$ & $\underline{\boldsymbol{0.955}}_{\pm 0.018}$ & $0.726_{\pm 0.002}$ & $0.919_{\pm 0.002}^*$ \\
\cline{1-10}
\multirow{2}{*}{bank} & MD & $0.947_{\pm 0.000}$ & $0.965_{\pm 0.002}$ & $0.891_{\pm 0.001}$ & $\underline{\boldsymbol{0.983}}_{\pm 0.001}$ & $0.915_{\pm 0.000}$ & $\boldsymbol{0.983}_{\pm 0.005}$ & $0.894_{\pm 0.003}$ & $0.966_{\pm 0.002}^*$ \\
 & CR & $0.885_{\pm 0.019}$ & $0.884_{\pm 0.003}^*$ & $0.870_{\pm 0.009}$ & $\boldsymbol{0.932}_{\pm 0.030}$ & $0.903_{\pm 0.000}$ & $\underline{\boldsymbol{0.967}}_{\pm 0.008}$ & $0.870_{\pm 0.011}$ & $0.877_{\pm 0.014}$ \\
\cline{1-10}
\multirow{2}{*}{breast} & MD & $\boldsymbol{0.854}_{\pm 0.005}$ & $0.775_{\pm 0.024}$ & $0.832_{\pm 0.019}$ & $0.838_{\pm 0.001}$ & $0.728_{\pm 0.000}$ & $0.813_{\pm 0.005}$ & $\underline{\boldsymbol{0.900}}_{\pm 0.013}$ & $0.833_{\pm 0.009}^*$ \\
 & CR & $\boldsymbol{0.767}_{\pm 0.006}$ & $0.616_{\pm 0.063}$ & $0.709_{\pm 0.015}$ & $0.767_{\pm 0.000}$ & $0.636_{\pm 0.000}$ & $0.709_{\pm 0.006}$ & $\underline{\boldsymbol{0.781}}_{\pm 0.010}$ & $0.747_{\pm 0.004}^*$ \\
\cline{1-10}
\multirow{2}{*}{credit} & MD & $\underline{\boldsymbol{0.971}}_{\pm 0.003}$ & $0.937_{\pm 0.013}$ & $0.944_{\pm 0.003}$ & $0.926_{\pm 0.007}$ & $0.932_{\pm 0.000}$ & $0.932_{\pm 0.007}$ & $0.931_{\pm 0.004}$ & $\boldsymbol{0.968}_{\pm 0.003}^*$ \\
 & CR & $\boldsymbol{0.925}_{\pm 0.004}$ & $0.862_{\pm 0.030}$ & $0.894_{\pm 0.005}$ & $0.864_{\pm 0.007}$ & $0.861_{\pm 0.000}$ & $0.862_{\pm 0.008}$ & $0.848_{\pm 0.004}$ & $\underline{\boldsymbol{0.934}}_{\pm 0.007}^*$ \\
\cline{1-10}
\multirow{2}{*}{diabetes} & MD & $0.898_{\pm 0.004}$ & $0.923_{\pm 0.006}$ & $0.793_{\pm 0.042}$ & $0.769_{\pm 0.017}$ & $0.881_{\pm 0.000}$ & $\underline{\boldsymbol{0.955}}_{\pm 0.007}$ & $0.824_{\pm 0.012}$ & $\boldsymbol{0.929}_{\pm 0.008}^*$ \\
 & CR & $\boldsymbol{0.949}_{\pm 0.007}$ & $0.898_{\pm 0.004}$ & $0.913_{\pm 0.001}$ & $0.875_{\pm 0.003}$ & $0.921_{\pm 0.000}$ & $\underline{\boldsymbol{0.964}}_{\pm 0.002}$ & $0.858_{\pm 0.008}$ & $0.946_{\pm 0.001}^*$ \\
\cline{1-10}
\multirow{2}{*}{iris} & MD & $\boldsymbol{0.893}_{\pm 0.011}$ & $0.822_{\pm 0.022}$ & $0.762_{\pm 0.040}$ & $0.553_{\pm 0.024}$ & $0.870_{\pm 0.000}$ & $\underline{\boldsymbol{0.893}}_{\pm 0.003}$ & $0.809_{\pm 0.004}$ & $0.872_{\pm 0.007}^*$ \\
 & CR & $0.897_{\pm 0.007}$ & $0.590_{\pm 0.019}$ & $0.793_{\pm 0.047}$ & $0.506_{\pm 0.036}$ & $\boldsymbol{0.899}_{\pm 0.000}$ & $\underline{\boldsymbol{0.917}}_{\pm 0.007}$ & $0.783_{\pm 0.017}$ & $0.874_{\pm 0.010}^*$ \\
\cline{1-10}
\multirow{2}{*}{qsar} & MD & $\underline{\boldsymbol{0.931}}_{\pm 0.004}$ & $0.917_{\pm 0.003}^*$ & $0.891_{\pm 0.001}$ & $0.602_{\pm 0.005}$ & $0.887_{\pm 0.000}$ & $\boldsymbol{0.930}_{\pm 0.005}$ & $0.873_{\pm 0.007}$ & $0.898_{\pm 0.001}$ \\
 & CR & $\boldsymbol{0.911}_{\pm 0.009}$ & $0.863_{\pm 0.007}$ & $0.887_{\pm 0.010}^*$ & $0.605_{\pm 0.001}$ & $0.855_{\pm 0.000}$ & $\underline{\boldsymbol{0.914}}_{\pm 0.009}$ & $0.849_{\pm 0.012}$ & $0.874_{\pm 0.006}$ \\
\cline{1-10}
\multirow{2}{*}{wdbc} & MD & $0.923_{\pm 0.003}$ & $0.861_{\pm 0.005}$ & $0.824_{\pm 0.031}$ & $0.395_{\pm 0.017}$ & $0.873_{\pm 0.000}$ & $\underline{\boldsymbol{0.946}}_{\pm 0.004}$ & $0.749_{\pm 0.026}$ & $\boldsymbol{0.928}_{\pm 0.005}^*$ \\
 & CR & $0.939_{\pm 0.001}$ & $0.866_{\pm 0.007}$ & $0.934_{\pm 0.003}$ & $0.772_{\pm 0.010}$ & $0.896_{\pm 0.000}$ & $\underline{\boldsymbol{0.976}}_{\pm 0.002}$ & $0.907_{\pm 0.017}$ & $\boldsymbol{0.951}_{\pm 0.002}^*$ \\
\midrule
\multirow{2}{*}{Avg.} & MD & \textbf{0.925} & 0.898 & 0.852 & 0.751 & 0.881 & \textbf{\underline{0.931}} & 0.860 & $0.906^*$\\
& CR & \textbf{0.904} & 0.825 & 0.857 & 0.783 & 0.867 & \textbf{\underline{0.919}} & 0.839 & $0.892^*$\\
\bottomrule
\end{tabular}

    }
\end{table}

Table~\ref{tab:fidelity-details} shows the raw scores of MD and CR for fidelity comparing different baseline models with TAEGAN. TAEGAN is not the best among all experimented models but generally maintains a comparable performance with the state-of-the-art models, and is the best GAN model.

\begin{table}[t]
    \centering
    \caption{Raw $p$-values of DCR values indicating risk of privacy leakage. Values smaller than 0.05 are risky, which are highlighted in red.}
    \label{tab:dcr-details}
    \resizebox{\linewidth}{!}{
    \begin{tabular}{l|ccccccc|c}
\toprule
 & ARF & CTAB+ & CTGAN & TDDPM & GReaT & TabSyn & TVAE & TAEGAN \\
\midrule
adult & $0.486_{\pm 0.077}$ & $0.691_{\pm 0.424}$ & $0.687_{\pm 0.272}$ & $0.440_{\pm 0.125}$ & \textcolor{red}{$0.000_{\pm 0.000}$} & $0.654_{\pm 0.149}$ & \textcolor{red}{$0.003_{\pm 0.004}$} & $0.938_{\pm 0.036}$ \\
bank & $0.999_{\pm 0.000}$ & $0.849_{\pm 0.093}$ & $0.996_{\pm 0.003}$ & $0.873_{\pm 0.077}$ & \textcolor{red}{$0.040_{\pm 0.000}$} & $0.938_{\pm 0.027}$ & $0.709_{\pm 0.377}$ & $0.729_{\pm 0.182}$ \\
breast & $0.071_{\pm 0.060}$ & $0.333_{\pm 0.468}$ & $0.595_{\pm 0.415}$ & \textcolor{red}{$0.039_{\pm 0.026}$} & \textcolor{red}{$0.005_{\pm 0.000}$} & $0.357_{\pm 0.256}$ & $1.000_{\pm 0.000}$ & $0.220_{\pm 0.116}$ \\
credit & $0.197_{\pm 0.068}$ & $0.161_{\pm 0.124}$ & $0.462_{\pm 0.131}$ & $0.725_{\pm 0.261}$ & $0.990_{\pm 0.000}$ & $0.689_{\pm 0.246}$ & $0.581_{\pm 0.194}$ & $0.442_{\pm 0.134}$ \\
diabetes & $0.698_{\pm 0.230}$ & $0.890_{\pm 0.036}$ & $0.835_{\pm 0.187}$ & $0.990_{\pm 0.005}$ & $0.971_{\pm 0.000}$ & $0.825_{\pm 0.093}$ & $0.909_{\pm 0.049}$ & $0.840_{\pm 0.176}$ \\
iris & $0.796_{\pm 0.127}$ & $0.970_{\pm 0.039}$ & $0.932_{\pm 0.032}$ & $0.972_{\pm 0.038}$ & $0.187_{\pm 0.000}$ & $0.765_{\pm 0.149}$ & $0.770_{\pm 0.118}$ & $0.677_{\pm 0.229}$ \\
qsar & $0.668_{\pm 0.231}$ & $0.546_{\pm 0.193}$ & $0.669_{\pm 0.186}$ & $0.485_{\pm 0.306}$ & $0.063_{\pm 0.000}$ & $0.664_{\pm 0.238}$ & $0.351_{\pm 0.160}$ & $0.865_{\pm 0.100}$ \\
wdbc & $0.821_{\pm 0.029}$ & $0.647_{\pm 0.283}$ & $0.909_{\pm 0.068}$ & $1.000_{\pm 0.000}$ & $0.591_{\pm 0.000}$ & $0.853_{\pm 0.096}$ & $0.307_{\pm 0.170}$ & $0.935_{\pm 0.036}$ \\
\midrule
\# vio. & 0 & 0 & 0 & 1 & 3 & 0 & 1 & 0\\
\bottomrule
\end{tabular}

    }
\end{table}

Table~\ref{tab:dcr-details} shows the raw $p$-values indicating the risk of privacy leakage computed based on DCR. TAEGAN and many good-performing baseline models do not suffer from outstanding privacy leakage risk.

\begin{table}[t]
    \centering
    \caption{Time used for training and generation respectively in seconds of different models on each dataset.}
    \label{tab:time-details}
    \resizebox{\linewidth}{!}{
    \begin{tabular}{ll|ccccccc|c}
\toprule
Dataset & Action & ARF & CTAB+ & CTGAN & TDDPM & GReaT & TabSyn & TVAE & TAEGAN \\
\midrule
\multirow{2}{*}{adult} & Train & $258.758_{\pm 55.030}$ & $787.379_{\pm 3.334}$ & $753.259_{\pm 104.080}$ & $285.153_{\pm 0.410}$ & $10397.040_{\pm 9.203}$ & $1199.823_{\pm 146.744}$ & $1111.927_{\pm 2.790}$ & $758.236_{\pm 1.879}$ \\
 & Generate & $41.029_{\pm 0.363}$ & $0.429_{\pm 0.001}$ & $0.112_{\pm 0.001}$ & $63.388_{\pm 0.083}$ & $138.023_{\pm 0.205}$ & $2.409_{\pm 0.010}$ & $0.357_{\pm 0.005}$ & $2.204_{\pm 0.079}$ \\
\cline{1-10}
\multirow{2}{*}{bank} & Train & $121.480_{\pm 39.164}$ & $737.314_{\pm 0.138}$ & $544.578_{\pm 62.697}$ & $252.016_{\pm 0.266}$ & $10295.769_{\pm 21.720}$ & $1865.473_{\pm 418.665}$ & $878.170_{\pm 122.998}$ & $832.752_{\pm 0.931}$ \\
 & Generate & $42.208_{\pm 5.408}$ & $1.126_{\pm 0.006}$ & $0.733_{\pm 0.001}$ & $55.038_{\pm 0.077}$ & $149.872_{\pm 0.104}$ & $3.009_{\pm 0.063}$ & $0.975_{\pm 0.003}$ & $2.782_{\pm 0.006}$ \\
\cline{1-10}
\multirow{2}{*}{breast} & Train & $0.585_{\pm 0.015}$ & $8.479_{\pm 0.026}$ & $13.571_{\pm 2.035}$ & $6.136_{\pm 0.099}$ & $153.352_{\pm 0.134}$ & $1037.126_{\pm 281.166}$ & $6.340_{\pm 0.925}$ & $37.559_{\pm 0.480}$ \\
 & Generate & $0.876_{\pm 0.013}$ & $0.056_{\pm 0.008}$ & $0.051_{\pm 0.000}$ & $3.123_{\pm 0.007}$ & $3.434_{\pm 0.016}$ & $0.249_{\pm 0.003}$ & $0.055_{\pm 0.000}$ & $0.090_{\pm 0.000}$ \\
\cline{1-10}
\multirow{2}{*}{credit} & Train & $2.262_{\pm 0.089}$ & $8.400_{\pm 0.035}$ & $32.463_{\pm 8.548}$ & $8.102_{\pm 0.215}$ & $277.555_{\pm 0.388}$ & $726.006_{\pm 67.662}$ & $13.246_{\pm 1.809}$ & $68.799_{\pm 1.766}$ \\
 & Generate & $2.198_{\pm 0.088}$ & $0.043_{\pm 0.000}$ & $0.035_{\pm 0.000}$ & $4.526_{\pm 0.007}$ & $5.845_{\pm 0.008}$ & $0.167_{\pm 0.003}$ & $0.041_{\pm 0.000}$ & $0.133_{\pm 0.001}$ \\
\cline{1-10}
\multirow{2}{*}{diabetes} & Train & $1.100_{\pm 0.094}$ & $8.524_{\pm 0.407}$ & $20.414_{\pm 2.831}$ & $5.799_{\pm 1.191}$ & $130.793_{\pm 0.077}$ & $758.140_{\pm 41.333}$ & $7.944_{\pm 1.272}$ & $54.576_{\pm 0.720}$ \\
 & Generate & $0.943_{\pm 0.047}$ & $0.047_{\pm 0.001}$ & $0.036_{\pm 0.000}$ & $2.329_{\pm 0.822}$ & $1.496_{\pm 0.008}$ & $0.139_{\pm 0.000}$ & $0.039_{\pm 0.000}$ & $0.096_{\pm 0.001}$ \\
\cline{1-10}
\multirow{2}{*}{iris} & Train & $0.153_{\pm 0.002}$ & $7.267_{\pm 0.021}$ & $6.538_{\pm 0.504}$ & $3.485_{\pm 0.210}$ & $24.975_{\pm 0.018}$ & $620.864_{\pm 35.603}$ & $2.811_{\pm 0.518}$ & $13.702_{\pm 0.040}$ \\
 & Generate & $0.690_{\pm 0.010}$ & $0.033_{\pm 0.006}$ & $0.016_{\pm 0.000}$ & $6.208_{\pm 2.124}$ & $0.360_{\pm 0.015}$ & $0.098_{\pm 0.002}$ & $0.016_{\pm 0.001}$ & $0.039_{\pm 0.000}$ \\
\cline{1-10}
\multirow{2}{*}{qsar} & Train & $18.992_{\pm 1.985}$ & $18.389_{\pm 0.217}$ & $64.331_{\pm 11.582}$ & $7.369_{\pm 0.156}$ & $566.658_{\pm 23.451}$ & $694.890_{\pm 76.489}$ & $23.013_{\pm 2.183}$ & $209.272_{\pm 3.010}$ \\
 & Generate & $2.842_{\pm 0.012}$ & $0.206_{\pm 0.006}$ & $0.192_{\pm 0.000}$ & $2.825_{\pm 0.826}$ & $46.712_{\pm 3.062}$ & $0.336_{\pm 0.005}$ & $0.203_{\pm 0.001}$ & $0.506_{\pm 0.070}$ \\
\cline{1-10}
\multirow{2}{*}{wdbc} & Train & $5.344_{\pm 0.860}$ & $10.795_{\pm 0.089}$ & $24.175_{\pm 7.507}$ & $4.249_{\pm 0.056}$ & $265.340_{\pm 2.411}$ & $631.546_{\pm 68.070}$ & $9.356_{\pm 1.236}$ & $109.288_{\pm 0.061}$ \\
 & Generate & $1.410_{\pm 0.007}$ & $0.159_{\pm 0.001}$ & $0.154_{\pm 0.002}$ & $1.934_{\pm 0.010}$ & $65.821_{\pm 1.106}$ & $0.276_{\pm 0.005}$ & $0.159_{\pm 0.000}$ & $0.302_{\pm 0.002}$ \\
\midrule
\multirow{2}{*}{Avg.} & Train & 51.933 & 187.517 & 178.590 & 67.330 & 2660.705 & 1023.474 & 226.858 & 254.823 \\
& Generate & 10.147 & 0.302 & 0.208 & 15.776 & 47.157 & 0.952 & 0.270 & 0.798 \\
\bottomrule
\end{tabular}

    }
\end{table}

Table~\ref{tab:time-details} shows the raw computation time on each dataset. Results show that GANs are generally more efficient.

\subsection{Ablation Study}
\label{app:exp:abl}

\begin{table}[t]
    \centering
    \caption{Results of ablation study on MLE.}
    \label{tab:abl-mle-details}
    \resizebox{\linewidth}{!}{
    \begin{tabular}{l|ccc|ccc|ccc|cccc}
\toprule
 & \multicolumn{3}{c|}{adult} & \multicolumn{3}{c|}{credit} & \multicolumn{3}{c|}{diabetes} & \multicolumn{4}{c}{Avg.} \\
ML & LG & RF & XGB & LG & RF & XGB & LG & RF & XGB & LG & RF & XGB & All \\
\midrule
TAEGAN & $0.907_{\pm 0.001}$ & $0.905_{\pm 0.002}$ & $0.909_{\pm 0.001}$ & $0.820_{\pm 0.015}$ & $0.819_{\pm 0.025}$ & $0.833_{\pm 0.014}$ & $0.874_{\pm 0.010}$ & $0.860_{\pm 0.021}$ & $0.855_{\pm 0.010}$ & 0.867 & 0.861 & 0.866 & 0.865 \\
\midrule
w/o warmup & $0.907_{\pm 0.001}$ & $0.905_{\pm 0.002}$ & $0.910_{\pm 0.001}$ & $0.814_{\pm 0.007}$ & $0.801_{\pm 0.016}$ & $0.821_{\pm 0.011}$ & $0.863_{\pm 0.016}$ & $0.837_{\pm 0.037}$ & $0.838_{\pm 0.029}$ & 0.861 & 0.848 & 0.857 & 0.855 \\
gen. in 2 steps & $0.907_{\pm 0.001}$ & $0.905_{\pm 0.001}$ & $0.910_{\pm 0.000}$ & $0.820_{\pm 0.015}$ & $0.818_{\pm 0.024}$ & $0.830_{\pm 0.015}$ & $0.874_{\pm 0.010}$ & $0.855_{\pm 0.022}$ & $0.854_{\pm 0.022}$ & 0.867 & 0.859 & 0.864 & 0.864 \\
w/o log. freq. & $0.906_{\pm 0.001}$ & $0.904_{\pm 0.002}$ & $0.910_{\pm 0.001}$ & $0.822_{\pm 0.023}$ & $0.817_{\pm 0.014}$ & $0.817_{\pm 0.025}$ & $0.876_{\pm 0.009}$ & $0.856_{\pm 0.012}$ & $0.851_{\pm 0.004}$ & 0.868 & 0.859 & 0.859 & 0.862 \\
$\lambda_1=\lambda_2=1$ & $0.905_{\pm 0.001}$ & $0.903_{\pm 0.001}$ & $0.909_{\pm 0.001}$ & $0.804_{\pm 0.008}$ & $0.809_{\pm 0.032}$ & $0.798_{\pm 0.042}$ & $0.870_{\pm 0.010}$ & $0.863_{\pm 0.014}$ & $0.850_{\pm 0.012}$ & 0.859 & 0.858 & 0.852 & 0.857 \\
w/o discrete noise & $0.907_{\pm 0.001}$ & $0.905_{\pm 0.002}$ & $0.910_{\pm 0.001}$ & $0.807_{\pm 0.014}$ & $0.816_{\pm 0.027}$ & $0.830_{\pm 0.025}$ & $0.876_{\pm 0.008}$ & $0.856_{\pm 0.006}$ & $0.853_{\pm 0.011}$ & 0.864 & 0.859 & 0.864 & 0.862 \\
w/o pac & $0.906_{\pm 0.001}$ & $0.904_{\pm 0.001}$ & $0.909_{\pm 0.001}$ & $0.816_{\pm 0.019}$ & $0.803_{\pm 0.022}$ & $0.828_{\pm 0.003}$ & $0.869_{\pm 0.014}$ & $0.854_{\pm 0.003}$ & $0.851_{\pm 0.003}$ & 0.864 & 0.854 & 0.862 & 0.860 \\
w/o info. loss & $0.907_{\pm 0.001}$ & $0.905_{\pm 0.001}$ & $0.910_{\pm 0.000}$ & $0.819_{\pm 0.014}$ & $0.821_{\pm 0.020}$ & $0.819_{\pm 0.029}$ & $0.874_{\pm 0.009}$ & $0.857_{\pm 0.016}$ & $0.855_{\pm 0.012}$ & 0.867 & 0.861 & 0.861 & 0.863 \\
w/o interac. loss & $0.907_{\pm 0.001}$ & $0.904_{\pm 0.003}$ & $0.909_{\pm 0.001}$ & $0.819_{\pm 0.016}$ & $0.819_{\pm 0.016}$ & $0.833_{\pm 0.017}$ & $0.873_{\pm 0.009}$ & $0.860_{\pm 0.019}$ & $0.852_{\pm 0.014}$ & 0.866 & 0.861 & 0.865 & 0.864 \\
\bottomrule
\end{tabular}

    }
\end{table}

Table~\ref{tab:abl-mle-details} shows the raw results of the ablation study on MLE. Original TAEGAN setting has the best score compared to all the other ablation settings, which are TAEGAN with one design component removed.  The advantage is more obvious with more complicated downstream task (XGB).

\begin{table}[t]
    \centering
    \caption{Results of ablation study on fidelity.}
    \label{tab:abl-fidelity-details}
    \resizebox{\linewidth}{!}{
    \begin{tabular}{l|cc|cc|cc|cc}
\toprule
 & \multicolumn{2}{c|}{adult} & \multicolumn{2}{c|}{credit} & \multicolumn{2}{c|}{diabetes} & \multicolumn{2}{c}{Avg.} \\
Metric & MD & CR & MD & CR & MD & CR & MD & CR \\
\midrule
TAEGAN & $0.958_{\pm 0.002}$ & $0.919_{\pm 0.002}$ & $0.968_{\pm 0.003}$ & $0.934_{\pm 0.007}$ & $0.929_{\pm 0.008}$ & $0.946_{\pm 0.001}$ & 0.952 & 0.933 \\
\midrule
w/o warmup & $0.959_{\pm 0.003}$ & $0.919_{\pm 0.005}$ & $0.969_{\pm 0.002}$ & $0.918_{\pm 0.006}$ & $0.933_{\pm 0.004}$ & $0.948_{\pm 0.003}$ & 0.954 & 0.928 \\
gen. in 2 steps & $0.958_{\pm 0.002}$ & $0.919_{\pm 0.002}$ & $0.968_{\pm 0.003}$ & $0.934_{\pm 0.007}$ & $0.929_{\pm 0.008}$ & $0.946_{\pm 0.001}$ & 0.952 & 0.933 \\
w/o log. freq. & $0.968_{\pm 0.000}$ & $0.932_{\pm 0.000}$ & $0.970_{\pm 0.004}$ & $0.935_{\pm 0.007}$ & $0.932_{\pm 0.007}$ & $0.948_{\pm 0.001}$ & 0.957 & 0.938 \\
$\lambda_1=\lambda_2=1$ & $0.963_{\pm 0.002}$ & $0.922_{\pm 0.003}$ & $0.967_{\pm 0.004}$ & $0.931_{\pm 0.007}$ & $0.931_{\pm 0.011}$ & $0.945_{\pm 0.002}$ & 0.953 & 0.933 \\
w/o discrete noise & $0.960_{\pm 0.001}$ & $0.920_{\pm 0.001}$ & $0.966_{\pm 0.004}$ & $0.922_{\pm 0.015}$ & $0.929_{\pm 0.008}$ & $0.945_{\pm 0.004}$ & 0.952 & 0.929 \\
w/o pac & $0.959_{\pm 0.001}$ & $0.919_{\pm 0.003}$ & $0.965_{\pm 0.002}$ & $0.913_{\pm 0.005}$ & $0.929_{\pm 0.001}$ & $0.946_{\pm 0.002}$ & 0.951 & 0.926 \\
w/o info. loss & $0.958_{\pm 0.001}$ & $0.918_{\pm 0.002}$ & $0.968_{\pm 0.004}$ & $0.934_{\pm 0.008}$ & $0.930_{\pm 0.008}$ & $0.946_{\pm 0.002}$ & 0.952 & 0.933 \\
w/o interac. loss & $0.958_{\pm 0.002}$ & $0.919_{\pm 0.002}$ & $0.968_{\pm 0.003}$ & $0.934_{\pm 0.007}$ & $0.929_{\pm 0.009}$ & $0.946_{\pm 0.002}$ & 0.952 & 0.933 \\
\bottomrule
\end{tabular}

    }
\end{table}

Table~\ref{tab:abl-fidelity-details} shows the raw results of the ablation study on fidelity metrics. Note that when uniform sampling instead of data sampling following logarithmic frequency is used, all fidelity metrics in all datasets are improved.

\end{document}